\documentclass[utf8]{FrontiersinHarvard}
\usepackage[utf8]{inputenc}
\usepackage{graphicx}
\usepackage{url}
\usepackage[dvipsnames]{xcolor}
\usepackage{caption}

\title[Legged Robots for Object Manipulation: A Review]{Legged Robots for Object Manipulation: A Review}

\author{kam37}
\date{January 2023}
\usepackage{url,hyperref,lineno,microtype,subcaption}
\usepackage[onehalfspacing]{setspace}
\usepackage{makecell}

\def\keyFont{\fontsize{8}{11}\helveticabold}
\def\firstAuthorLast{Gong and Sun {et~al.}} % use et al only if is more than 1 author
\def\Authors{Yifeng Gong$^{1,\dagger}$, Ge Sun\,$^{2,\dagger}$, Aditya Nair\,$^3$, Aditya Bidwai\,$^2$, Raghuram CS\,$^4$, John Grezmak\,$^1$, Guillaume Sartoretti\,$^2$, and Kathryn A. Daltorio\,$^{1,*}$}
\def\Address{
$^{1}$ Department of Mechanical Engineering, Case Western Reserve University, Ohio, USA.\\
$^{2}$ Department of Mechanical Engineering, National University of Singapore, Republic of Singapore.\\
$^{3}$ Department of Mechanical Engineering, Birla Institute of Technology and Science Pilani, India.\\
$^{4}$ Department of Electrical and Electronics Engineering, Birla Institute of Technology and Science Pilani, India.\\
$\dagger$ These authors contributed equally to this work and share the first authorship.}

\def\corrAuthor{Kathryn A Daltorio, Case Western Reserve University, 10900 Euclid Ave, Cleveland, Ohio 44106-7222}
\def\corrEmail{kam37@case.edu}

\begin{document}

\onecolumn
\firstpage{1}

\author[\firstAuthorLast]{\Authors} % This field will be automatically populated
\address{} % This field will be automatically populated
\correspondance{} % This field will be automatically populated
\extraAuth{} % If there are more than 1 corresponding author, comment this line and uncomment the next one.

\maketitle

\begin{abstract}
\section{}
Legged robots can have a unique role in manipulating objects in dynamic, human-centric, or otherwise inaccessible environments. 
Although most legged robotics research to date typically focuses on traversing these challenging environments, many legged platform demonstrations have also included ``moving an object'' as a way of doing tangible work.
Legged robots can be designed to manipulate a particular type of object (e.g., a cardboard box, a soccer ball, or a larger piece of furniture), by themselves or collaboratively.
The objective of this review is to collect and learn from these examples, to both organize the work done so far in the community and highlight interesting open avenues for future work.
This review categorizes existing works into four main manipulation methods: object interactions without grasping, manipulation with walking legs, dedicated non-locomotive arms, and legged teams.
Each method has different design and autonomy features, which are illustrated by available examples in the literature.
Based on a few simplifying assumptions, we further provide quantitative comparisons for the range of possible relative sizes of the manipulated object with respect to the robot.
Taken together, these examples suggest new directions for research in legged robot manipulation, such as multifunctional limbs, terrain modeling, or learning-based control, to support a number of new deployments in challenging indoor/outdoor scenarios in warehouses/construction sites, preserved natural areas, and especially for home robotics.

\tiny
 \keyFont{ \section{Keywords:}
Multilegged Robots;
Mobile Manipulation;
Grasping;
Legs as Grippers and Other End-Effectors;
Cooperative Manipulation;
Mechanism Design of Mobile Robots;
Domestic Robots;
Field Robots;
}
\end{abstract}

\section{Introduction}

Research on legged robots design and locomotion has mainly been fueled by the desire to deploy teleoperated or autonomous systems in otherwise inaccessible terrains.
While wheeled and tracked vehicles are broadly used on paved surfaces (currently no more than 7$\%$ of Earth's land surface) and fields for agriculture and forestry (roughly 46.6$\%$)~\citep{LandReview}, they have poor performance on sandy, rocky, and other unmodified natural terrains (roughly 46.5$\%$). 
Nearly half of Earth's ice-free land area has not been modified by humans~\citep{LandReview}, and is often inhabited by animals that use their legs to survive by
walking~\citep{1308761}, running~\citep{Bigdog}, climbing~\citep{728488}, jumping~\citep{zhang2017survey}, and swimming~\citep{song2016turtle}.
In addition to accessing natural terrains, legs can also enable improved access to human environments by climbing stairs, changing body height to crawl through confined spaces, or carefully stepping around clutter or hazards.
Potential real world applications such as industrial inspection can also rely on legged robots to regularly validate visual, thermal, and acoustic data at waypoints in monitoring and exploration tasks (see recent relevant review~\cite{bellicoso2018advances}).
Furthermore, legged robots can deliver payloads such as medkits in search-and-rescue scenarios to hard-to-reach locations, e.g., after standard transport routes have been compromised by a natural disaster.

Impressive recent advances across different fields of robotics now bring us closer to future legged robots that more actively interact with new environments.
Legged robots can learn to traverse challenging or urban environments~\citep{miki2022learning,UnderwaterManipulation}, by adapting their gait, footholds, and body pose, and the next frontier lies beyond passive exploration (such as mapping/inspection/localization).
Actively turning a valve~\citep{zhao2019novel}, retrieving an object~\citep{roennau2014lauron}, pushing away an obstacle~\citep{stuber2020let} or opening a door~\citep{schmid2008opening} are some fundamental manipulation tasks that can be the difference between mission success and failure for a legged robot, and can drastically enhance their abilities.
Fortunately, decades of manipulation work has developed an ecosystem of platforms capable of both fine dexterous in-hand-manipulation (e.g., robotic Rubik's cube solver~\citep{yang2020benchmarking}) and inexpensive, high-volume pick and place robots~\citep{altuzarra2011symmetric} with ongoing goals to approach human-like manual dexterity~\citep{billard2019trends}.
Manipulation research has primarily focused on stationary arms, but manipulation also occurs on mobile platforms (see relevant reviews on wheeled robots~\citep{thakar2023survey} and on RoboCup~\citep{sereinig2020review}).
Part of the reason why legged robots have often remained limited to simple interactions with their surroundings is that they are usually deployed in environments that are less predictable, making legged manipulation tasks more challenging.
However, some of the most advanced legged robots are overcoming these challenges to perform dynamic manipulation (e.g., catching a ball, or opening a door) during locomotion~\citep{zimmermann2021go,bellicoso2019alma}.
These works demonstrate the promising potential of legged manipulation platforms as one of the next frontiers for the legged robot community.

In this review, we consider different approaches to moving objects with a legged platform.
While ``manipulation'' encompasses any robotic efforts to change the state of an object~\citep{lynch_park}, a unique ability for mobile robotic platforms is changing the position of an object.
Focusing on moving objects also enables us to compare fundamentals of the relationships between an independent object to be moved and different legged robotic platforms.
These concepts apply broadly to applications in factories, hospitals, agriculture, land management, and domestic environments.
A wide variety of other potentially useful examples of interactions such as digging in granular media, adjusting a component of a larger structure, pressing buttons, or driving a car are more specialized, but often include some of the same primitive abilities.

The approach selected for manipulation also determines the overall design of the robot and system.
If a legged platform is carrying task-specific manipulators, design metrics such as weight, range, strength, and generality are important.
If the legs themselves become tools for moving objects, key design trade-offs become impact resistance and sensitivity.
Teams of robots can also collaborate to move larger objects together, but each robot then is typically smaller and more dependent on coordination.
In all cases, the stability of the platform must be considered to prevent robots from falling over and ensure useful contacts with the environment are leveraged, while the robustness of the grasp must be considered to prevent objects from being dropped or damaged.

While legged robots have made significant strides in manipulation capability in recent years, a proper summary and review of the current progress in legged robot manipulation is missing.
Most reviews surrounding legged robots focus on mechanical designs and control strategies that power their impressive legged locomotion ability~\citep{zhou2012survey,aoi2017adaptive,silva2012literature,suzumori2018trends,liu2007legged,sayyad2007single,wu2009survey}.
However, many recent works in the community have started focusing on legged robot manipulation, and this upcoming field at the interface between legged robot and (mobile) manipulation is becoming popular and active research area, which has not been covered by any other review paper to our knowledge.
This work aims to fill this gap by providing an in-depth analysis of the various types of legged robot manipulation.
Specifically, our aim is to present a comprehensive summary of the general challenges, design, control strategies, and features of legged robot manipulation, as well as to highlight rising opportunities for future research.
We also hope that this work can assist new researchers in this field, or those coming into it from an interdisciplinary background, in selecting and designing new legged robots and approaches for legged manipulation tasks.

The outline of this review is as follows: first, we provide a brief background on general requirements related to legged robot manipulation.
We then classify manipulation approaches into four main categories: (a). Object interactions without grasping (Pushing, kicking, and non-prehensile lifting). (b). Manipulation with walking legs (single leg, double leg, and whole body). (c). Manipulation with dedicated non-locomotive arms (e.g., when an arm or trunk is added to the legged platform). (d). Legged teams for manipulation (when a legged robot helps other robots or humans with an object interaction).  
We propose a new theoretical analysis to relate the size of manipulated objects and the manipulation limits of robots that use legs to grasp (fig.~\ref{fig:GraspAnalysis_Results}).
Through a number of new figures and tables, we summarize the carrying capacity of the different manipulation methods discussed in this work, by comparing the maximum object size v.s. maximum robot size.
In doing so, we examine the key characteristics of different manipulation methods, with the goal of helping future users select the proper system and control approach based on the required tasks.
The examples provided, most of which have been validated on physical platforms.
We finally discuss open avenues for future research, including both open/new challenges in this area and future applications for these platforms.

\begin{figure}[h!]
\begin{center}
\includegraphics[width = \textwidth]{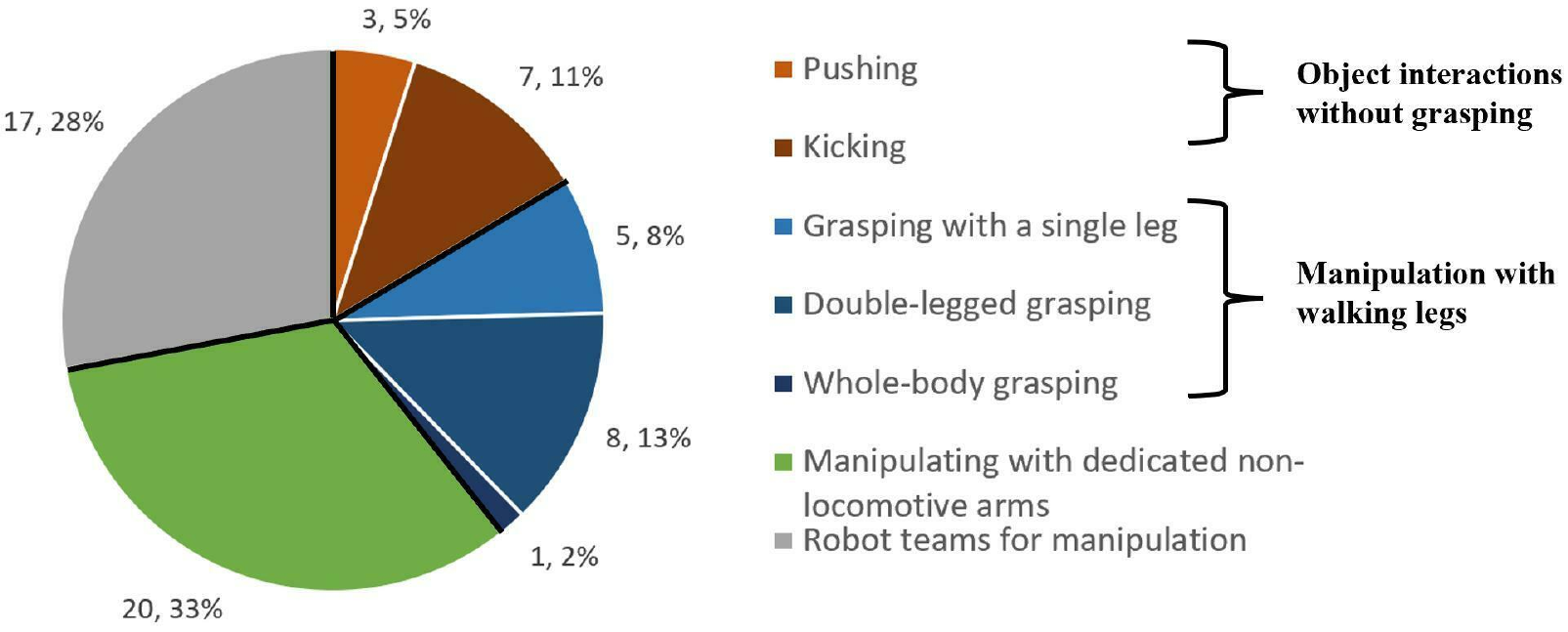}
\end{center}
\caption{Our work categorizes legged platforms for object manipulation into four main groups based on grasp type: Object interactions without grasping, Manipulation with the walking legs, Dedicated Non-Locomotive Arms, and Legged Teams. Sixty-one papers have been validated with physical robots. As an overview, this pie chart shows that the number of publications is roughly evenly divided between these types. 
}\label{fig:Category}
\end{figure}

\section{General Requirements}

This section discusses the general requirements/challenges that all types of legged manipulation platforms face.
First of all, we note that legged robot manipulation is much more complex than pure locomotion or pure manipulation, even though each of these two parts bring their own challenges.
However, since the robot's body, legs, and manipulated object also affect each other, one of the key new challenges lies in ensuring simultaneous stabilization of both locomotion and manipulation, even during dynamic configuration changes in complex environments.

First, a legged platform's stance must be stable with respect to the environment during manipulation.
In other words, the robot should not be destabilized by manipulation motions or reaction forces.
To maintain the static stability of a legged robot, the projected center of mass (along the gravity vector) must fall within its support polygon (i.e., the convex hull constructed from all ground contacts, i.e., leg positions)~\citep{mcghee1968stability}.
When a legged robot extends its manipulator to reach an object, the weight of the manipulator and the object may move the robot's CoM out of its support polygon.
This can explain why grasping an object is harder for bipedal robots, since their support polygons are relatively small, only formed by two legs, and the mass of the torso and arms can easily shift the CoM outside of such a small area~\citep{kim2021bipedal}.
Adding more legs can help mitigate this problem, thus fueling increasing research in quadruped- and hexapod-based manipulation~\citep{van2022collision,xin2022loco,yao2022transferable,kalouche2015modularity,fang2018type}. 
Conversely, we note that pushing against a large object can actually increase the robot's stability, if the object can help support some of the robot's weight.

Second, contact with the object to be manipulated must be reliable and safe, to prevent the robot from dropping or damaging the object.
The different classes of legged manipulation approaches described in this work exhibit different types of contact forces.
For most types of grasping discussed, the ultimate goal is to lift the object off the ground, implying the need for either \textit{form closure}~\citep{prattichizzo2016grasping} (i.e.,  kinematically constraining a rigid body by stationary contacts) or \textit{force closure} (i.e., securing objects by frictional contacts).
However, non-grasping (non-prehensile) manipulation is another way of moving objects without the requiring a secure grasp, by instead applying forces to push, impacts to kick, or relying on another (non-grasping) type of attachment mechanism to the object such as hooks/trays~\citep{ruggiero2018nonprehensile}.
Performing manipulation tasks with a legged platform also increases the potential vibrations transmitted to the manipulator, which may need to be corrected/accounted for to maintain continuous and secure contact with the object \citep{ferrolho2022roloma}. 

The third and final requirement is related to the feedback effects between the locomotor and manipulator, with the aim to minimize the effects that the manipulator can have on the robot's stability that can affect manipulation accuracy.
In manufacturing, industrial arms are usually installed on a fixed base so that end-effectors' positions and rotation angles can be easily and accurately calculated by Denavit Hartenberg transformation~\citep{DH} or Product of Exponentials~\citep{POE}.
For mobile manipulation, manipulators have been installed on wheeled platforms for pick-and-place tasks; but there, relatively precise kinematic and dynamic models still exist, which can even take into account slip~\citep{liu2009modeling}.
Although legged robots can generate precise gait patterns without payload, increasing the mass carried by the robot can dramatically decreases locomotion/stability accuracy and reliability.
While elasticity (i.e., \textit{compliance}) in the robot's legs may enables smoother suspension of loads, these can also further decrease control/motion accuracy~\citep{1248924}. 
In addition, granular media, like sand or pebbles, are often found in natural terrains and are hard to model due to their complexity and stochastic nature~\citep{nagabandi2018learning}.
Keeping the robot's body stable on these surfaces is thus even more challenging, because ground penetration varies based on the weight supported by each leg, which changes as objects are moved/lifted/dropped or gait cycles modified.
More generally, we note that legged manipulation systems are typically hybrid, underactuated, and nonlinear, making the development of controllers for legged-base manipulators complex, especially in the presence of additional manipulation appendages (manipulating with non-locomotive arms)~\citep{sleiman2021unified}.
Potential solutions can involve continuous feedback to correct relative motion errors, coordinate between locomotor and manipulator to improve the overall accuracy (i.e., \textit{whole-body control}), as well as adaptive controllers that learn and improve over time.

\section {Types of Manipulation }
\label{Types of Manipulation}

In this section, we discuss each of the four main categories of legged robot manipulation. For each category, we bring up the fundamental methodology, system design, and control strategy of existing works, as well as the pros and cons of each strategy. All cited papers that were validated with physical robots are also summarized in Fig.~\ref{fig:Category}. We further summarize the carrying capacity and control strategy of the different manipulation methods discussed in this section in Fig.~\ref{fig:GraspingCapacity} and Table.~\ref{Table of legged robots and control methods} respectively.

\subsection{Object Interactions without Grasping}
\subsubsection{Locomoting into an object: Pushing}
\label{Pushing}

Conceptually, pushing an object is one of the simplest methods for a legged robot to manipulate its environment. 
Here, we define pushing as walking such that the robot's body contacts an object imparting some horizontal force, such that the object slides along the ground.
Any mobile robot can move an object by locomoting into it, provided it can impart enough pushing force.
To successfully move an object, the pushing force must exceed the resistance force, which can be approximated with Coulomb friction on flat surfaces:
\begin{equation}
    F_{push} \geq \mu W,
\end{equation}
\noindent where $F_{push}$ is the force applied by the robot, $\mu$ is the coefficient of friction, and $W$ is the weight of the object to be moved.

However, the effect of the pushing force on the object depends on the location of the pushing force, the object's weight distribution, and shape, as well as its contact(s) with the ground.
Robots can easily (and inadvertently) push over objects~\citep{5509220}, navigating an obstacle course of tall thin objects without pushing them over is an effective demonstration of agility.
To push while avoiding knocking over the object, the moment caused by the pushing force must be less than the moment of the object's weight about the tipping edge.
Specifically, for a rigid object,
\begin{equation}
    F_{push} h \leq W x_{CoM},
\end{equation}
\noindent where $h$ is the height of the point of application of force, and $x_{CoM}$ is the horizontal distance of the object's center of mass from the tipping edge.
Thus, a pushing strategy is most likely to be effective in moving relatively light and wide objects on flat smooth surfaces rather than heavy and tall objects on rough terrains. 

The mechanics of pushing helps inform the hardware design of robots for these environments.
Because of Newton's third law, equal and opposite reaction forces will also be applied to the robot during pushing that can cause the robot to similarly slip or tip, which means the configuration of the robot is similarly constrained. 
As a result, pushing robots tend to be relatively heavy, wide-based, and powerful.
In addition, broad contact surfaces may reduce the precision of the contact calculations, but can help provide evenly distributed forces.
These design concepts are evident in robot snowplows~\citep{klein2016autonomous}; their wide blades provide an even force distribution and extra weight is added to generate higher traction forces.
However, wheeled robots are limited to flat terrains. 
The additional degrees of freedom offered by legged robots can enable locomotion over a larger variety of structured and unstructured challenging environments.
A robot might climb into a truck, climb up stairs, walk over rugged terrain, or walk behind an object to begin pushing. 

A second advantage of pushing with robotic legs is that the legs typically have more strength and provide more range of motion than the upper appendages alone.
The loading capacity of arms is enhanced by actuating the whole body.
By erasing the boundary between walking and working spaces, the overall workspace is expanded in both size and dimension.
As a result, the forces applied to the object can be much higher and the load on the actuators is better distributed.
In~\cite{1013568} and~\cite{936465}, a humanoid robot changes its body's configuration by adjusting its posture and changing its footholds. 
There, an objective function based on the \textit{manipulability} (which essentially measures how easily the robot can instantaneously move in any direction, as defined in~\cite{Lynch2017ModernRM}) of both its arms is maximized, making it more dexterous.
\cite{1013570} uses a similar approach to optimize the robot's body posture to uniformly distribute load across the robot's actuators, comparable to proper lifting techniques for humans.
This also applies to pushing as well to other scenarios that use dedicated non-locomotive arms as discussed in Section~\ref{Dedicated arm}.

Multi-legged robots can lean on and rest some of their legs on an object to push it more stably with their remaining legs.
For example,~\cite{5509220} utilizes a hexapod robot, which leans against a large block while holding it with two legs to push it along the floor.
Leaning while holding establishes three points of contact between the robot and the object, equivalent to secure force closure, as discussed in Section~\ref{Multi-Legged-Manipulation}.
Furthermore, leaning provides stability to the robot due to the wider resulting base of support.
Finally, leaning puts some of the robot's weight onto the pushed object, increasing and stabilizing the available contact force.
Since the robot is supported by the object, the remaining four legs can work in pairs to provide bilateral symmetric pushing forces (i.e., the middle left and right legs push together while the back left and right legs swing and vice versa).
These gaits can be optimized for more stable pushing, bypassing much of the complicated friction modelling.
With more complex modeling, the leg could even specifically brace against particular footholds to exploit higher effective coefficients of friction.

In cases where the robot makes and breaks contact with the object, model-based control has been often used for its inherent robustness and safety guarantees.
Unlike simple open-loop cases, where the robot is assumed to remain rigidly fixed to the object, closed-loop pushing addresses the complex contact dynamics incorporating friction between the robot and the object.
Most existing approaches to this problem as well as other examples of non-prehensile manipulation with legged robots, like kicking, rely on model-based and model-free hierarchical approaches.
These employ optimization methods such as Model Predictive Control (MPC) or Reinforcement Learning (RL), often using a high-level planner on top of a low-level controller.
For example,~\cite{rigo2022contact} aims to minimize tracking error when tasking a quadruped robot by pushing a large box using its body along a given trajectory on the floor.
Similarly,~\cite{sombolestan2022hierarchical} considers a quadruped robot tasked with pushing a large and heavy box (up to 60\% of the robot's weight) up a slope with no knowledge of the object's mass, coefficient of friction, or slope of the terrain.
In both of these methods, separate mode sequences are established for frictional contact, while a real-time controller is used to connect these contact modes into a feasible plan.
This is done through two cascading MPCs where the high-level manipulation MPC optimizes contact forces and location while the low-level ``loco-manipulation'' MPC regulates interaction forces between the robot and the object during the locomotion task.

A fundamental limitation of pushing is that the object's friction with the ground must be overcome.
One way effective friction can be reduced is if the object can be rolled.
Although only validated in simulation, an interesting example is the dual MPC-based approach described in~\cite{9341218}, where, much like a dung beetle, a quadruped robot stands on top of a large ball and rolls it to transport itself and the ball.
The other way to avoid losing energy and momentum to friction is to lift the object, as discussed in subsequent sections.

\subsubsection{Imparting an impulse: Kicking}
Kicking imparts a quick impulse rather than a sustained force on the object.
In other words, some of the robot's momentum is imparted to the object during their (usually brief) contact.
This is typically treated differently since there is no attempt to maintain or continually reestablish contact with the object.
The object to be manipulated is ideally lightweight to facilitate projectile motion, elastically deformable to minimize any damage due to large impulses, and round to facilitate rolling as well as mitigate any aerodynamic, structural, or contact considerations due to shape.
This is why balls made of flexible materials are well-suited and the most common candidates.

The problem of robot kicking using legs has broadly been tackled using two distinct approaches. 
The first is to develop a robot that can kick a ball a certain large distance accurately while the robot itself remains stationary. 
This approach deals with the problem of striking a ball but ignores that of locomoting and maintaining stability while kicking.
\cite{rober2021worlds} demonstrated a stationary kicking robot using springs that could kick an American football up to 64m without breaking itself.
However, the speed and point-of-contact during kicking were not precise resulting in a large variance in the ballistic trajectory of the ball and hence, the landing location.
\cite{flemmer2014humanoid} constructed a similar kicking robot powered by pneumatic cylinders with an automatic vision-based aiming system using an onboard camera.
This robot could consistently kick a ball for an average distance of 45.6m.
\cite{shankhdhar2022dynamics} made a mobile kicking robot by mounting a kicking apparatus onto a holonomic wheeled base.
Although this method allows a mobile robot to kick objects, it may not be considered a true kicking legged robot since the kicking ``leg'' is purely for striking and therefore non-locomotive.
More recent attempts go a step further and model the rigid contact dynamics by devising model-based and model-free RL approaches for stationary robots tasked with striking a hockey puck~\citep{chebotar2017combining} and playing mini golf~\citep{Muratore2021NeuralPD} respectively.

The second approach is to use walking robots to additionally kick/dribble while locomoting.
The most well-known proponents for this research have been the RoboCup Leagues~\citep{stone2007intelligent}.
The RoboCup Leagues consist of various leagues primarily centered around robot soccer, but also include search-and-rescue competitions using legged robots.
Robot soccer leagues are simulation or hardware competitions divided into three classes: quadruped robots, medium-sized humanoids, and small-sized humanoids.
In the quadruped class, only Sony's fully autonomous robot dogs, AIBO, are allowed, while in the humanoid class, each robot must have two legs, two arms, a head, and a trunk within specific restrictions on the height of the center of mass and size of the feet.
This has inspired a large body of work based on optimizing robot vision, robotic learning, localization, and locomotion strategies dedicated to these well-defined problem statements.
In most early works such as~\cite{cherubini2010policy, friedmann2008versatile, behnke2008hierarchical, acosta2008modular}, the task of kicking a ball into a goal is broken up into several simpler behaviours such as approaching, controlling, and reaching, with further-distinguished sub-behaviours such as kicking to the front or the side, etc. (which is always carried out using their heads to control and kick the ball).
Though only validated in simulation,~\cite{Jouandeau2014OptimizationOP} is one of the first attempts to improve shooting distance using trajectory optimization, while~\cite{peng2017deeploco, da2021deep, Teixeira2020HumanoidRK} developed an RL based approach to teach humanoid robots to dribble, kick, and shoot. 

Quadrupeds can also accurately kick a ball into a goal~\citep{Ji2022HierarchicalRL}.
This is an example of a contact-rich task, where MPC is not practical since the dynamics at play are difficult to model accurately (e.g., deformable contacts with the ball).
Furthermore, even if an accurate model could be obtained, it would still be computationally expensive and slow, thus unsuitable for such an online application.
Thus, in that work, hierarchical RL is used with a high-level trajectory behaviors policy on top of a low-level kicking motion control policy.
The low-level control policy swings the leg quickly while keeping the robot stable, and the high-level planning policy approximates the contact between the robot and the ball, the rolling friction, and the interaction of the ball with the world for sending the ball to the goal.
Multi-stage training is performed with a rigid ball in simulation and fine-tuned with a soft ball in the real world, toward a final demonstration on hardware.

\subsubsection{Non-prehensile Lifting}

Finally, it is also possible to use legged robots for other types of non-prehensile lifting, although few examples exist yet in the literature.
For example, an object and body robot could have mating attachment points (such as a hook and an eye), which allows the robot to lift the object off the ground and carry it without any grasping involved.
A robot waiter could have a long flat protrusion to slide under a tray (similar to the way a forklift carries pallets.) A robot might have an electromagnet to collect metals from a field.
For more examples, a rigorous review of non-prehensile manipulation and manipulation primitives is provided in~\cite{ruggiero2018nonprehensile} and applied to a wheeled robot in~\cite{bertoncelli2020linear}.   
The legs of the platform must be able to support the additional weight of the object or objects. If the center of mass of the robot changes, walking gaits may need to be adapted.
Legged platforms will induce more oscillation than wheeled robots, which may require them to choose slower speeds or more secure attachments, such as those offered by the grasping approaches described in the next sections.

\subsection{Manipulation With Walking Legs}
\label{Manipulation With Walking Legs}

\begin{figure}[h!]
\begin{center}
\includegraphics[width = \textwidth]{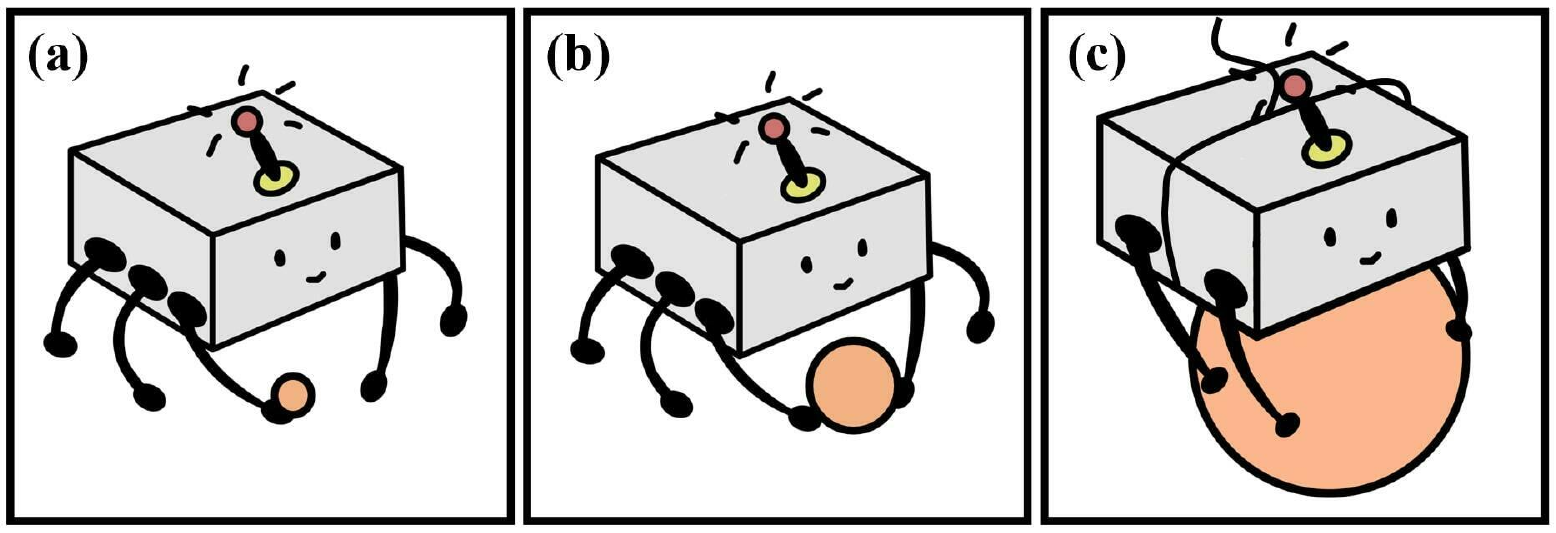}
\end{center}
\caption{Main classes of grasping with walking legs (left to right): (a) grasping with a single leg~\citep{heppner2015laurope}, (b) grasping with two legs~\citep{booysen2015gait}, and (c) grasping using all 
 legs/whole body~\citep{YifengGong}. Image by and courtesy of Yixian Qiu.}
\label{fig:legs}
\end{figure}

In robot designs where legs have many degrees of freedom for locomotion, these same degrees of freedom can also be suitable for directly grasping and manipulating objects.
We classified manipulation with walking legs into three categories based on the number of legs used (Fig.~\ref{fig:legs}): Grasping with a single leg, multi-legged grasping (usually two), and whole-body grasping, in which no legs are reserved for walking on the ground.

\begin{figure}
\begin{center}
\includegraphics[width = \textwidth]{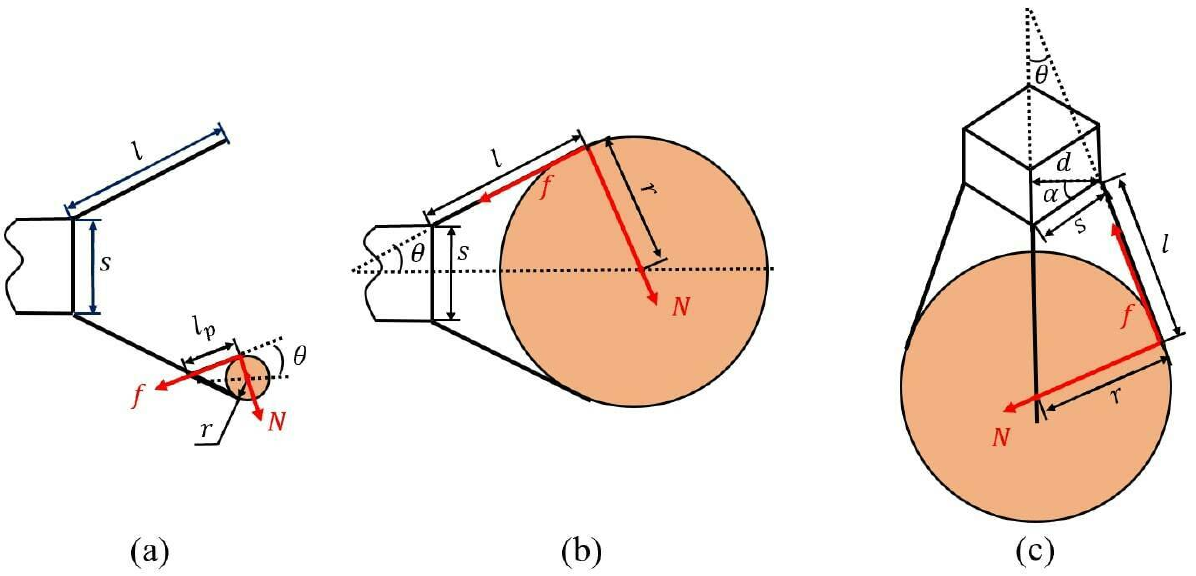}
\end{center}
\caption{Orthogonal top views of (a) one-legged and (b) double-legged grasping.  Isometric view of (c) whole-body grasping for a quadruped.}  
\label{fig:grasp_view}
\end{figure}

We analyze the relationship between the size of the manipulated object (relative to the robot) and the manipulation challenges/limitations of these three methods of manipulation. 
The required coefficient of contact friction $\mu$ represents the manipulation difficulty.
For a given value of $\mu$, each manipulation method has a different maximum grasp size to ensure \textit{force closure}.
We derive these limits for each manipulation method by assuming the object to be a sphere for generality, thin legs for simplicity, and non-backdrivable actuators with no torque limits as this analysis is limited to spatial considerations for the design parameters of the robot. 
The object radius is $r$, the fully extended leg length is $l$, the distance between the shoulders of two legs (width) is $s$, and the pincer length is $l_{p}$.
The dimensions of our models are based on LAURON V~\citep{roennau2014lauron}: $l = 2\, s = 4\, l_{p}$ while the grippers are assumed to be scissor-like pincers.
By forming geometric models based on Coulomb friction, we obtain the following constraints for \textit{force-closure}.

\subsubsection{Grasping with a Single Leg}
\label{Single leg}

\begin{figure}[htbp]
\begin{center}
\includegraphics[width = \textwidth]{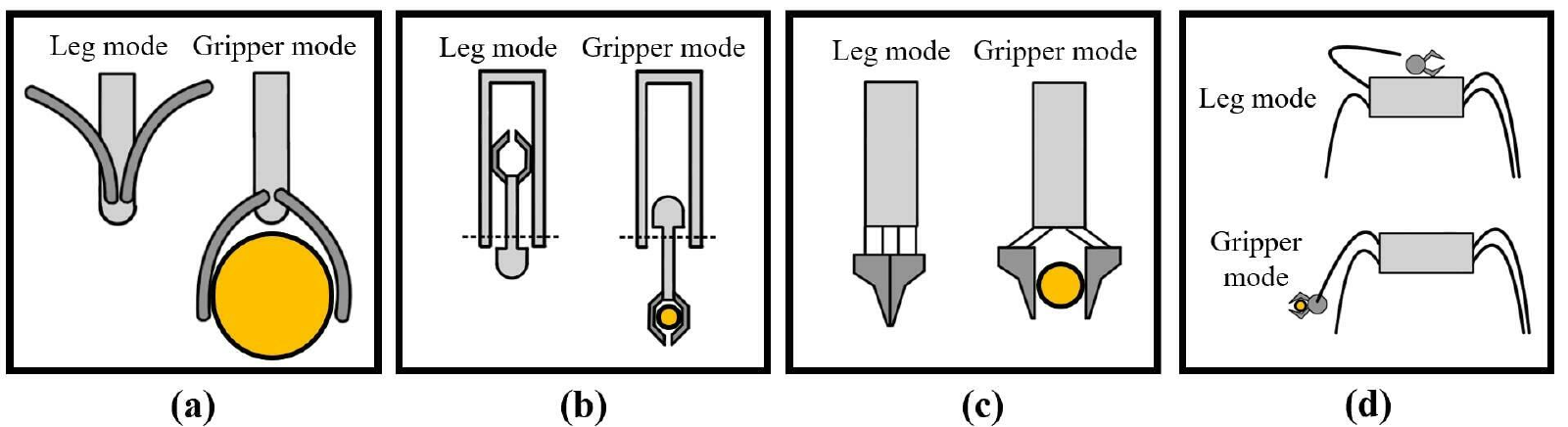}
\end{center}
\caption{Designs of individual leg manipulators: (a)~\cite{heppner2015laurope}, (b)~\cite{kim2014design}, (c)~\cite{Reconfigurable}, (d)~\cite{hirose2005quadruped}. While the first three designs illustrate integrated gripper-to-leg designs, the last design features grippers as modular tools.
}\label{Legdesign}
\end{figure}

For grasping with a single leg, the robot's gripper design is heavily dependent on the design of its legs (Fig.~\ref{Legdesign}). 
\cite{heppner2015laurope} presented a bio-inspired hexapod, LAURON V, which includes a lightweight gripper on one front leg that can be retracted actively when needed (e.g., during locomotion on rough terrain).
This gripper-to-leg design (Fig.~\ref{Legdesign}a) is necessary for the robot to protect the gripper from directly supporting the robot's heavy body. 
By shifting its CoM backward, LAURON V can stand on its four hind legs, leaving the two front legs free for manipulation.
To prevent stability loss when the robot walks with a grasped object, LAURON V stores the object on its back so that the front legs can be freed to support and push the body forward during locomotion.
Similarly,~\cite{kim2014design} developed a hexapod for underwater locomotion featuring two front grippers which can be concealed inside the robot's shins during walking (Fig.~\ref{Legdesign}b). 
Compared with LAURON V, these grippers each possess three DoFs but are limited in size by the shins they are retracted into. 
\cite{Reconfigurable} designed a scissor-like gripper-to-leg for the hexapod Hexaquad, that two scissor tips can be directly converted into leg mode by closing tips together or into gripper mode by splitting them (Fig.~\ref{Legdesign}c).
While many legged robots only use front legs for manipulation, Hexaquad expands the workspace since all legs are gripper-to-legs that can be transformed to be grippers whenever needed. In other words, Hexaquad does not need to turn around to grab an object that is behind it. To ensure a large support polygon for both front leg and back leg manipulation, the base joints of the middle legs are capable of moving back and forth.
For certain applications, designing grippers as interchangeable tools can be essential for improving versatility (Fig.~\ref{Legdesign}d).
The quadruped TITAN-IX can store tools (including grippers) on its back and then equip them on its end-effectors for demining tasks~\citep{hirose2005quadruped}. 
When retrieving an object, one leg works as a manipulator, and the other three legs keep the body stable. 
Developing methods for controlling such manipulators is a notable challenge, especially when performing such contact-rich tasks.
TITAN-IX implements a human-robot interface for cooperative manipulation using a Leader-Follower Manipulator: As a human operator controls a master gripper through haptic feedback, the robotic gripper tracks the operator's motion.

For grasping with a single leg (Fig.~\ref{fig:legs}a and Fig.~\ref{fig:grasp_view}a), two pincers grip the object with the friction generated due to pressing against it. The friction force component acting towards the gripper must be greater than the normal force component squeezing the object away i.e., $\mu N cos(\theta)> N sin(\theta)$, or  $\mu > tan(\theta)$. 
Here $\theta$ is half of the angle subtended by the pincers. Therefore:
\begin{equation}\label{Individual_Leg_eq}
\mu > \frac{r}{l_{p}} = 4 \cdot \frac{r}{l}
\end{equation}
where pincer length $l_p$ can be expressed as $l_p = \frac{l}{4}$.

\subsubsection{Multi-legged Grasping}
\label{Multi-Legged-Manipulation}

Multiple legs working together can lift larger objects better than a single gripper can, owing to a larger grasp and distributed load. 
\cite{gong_grezmak_zhou_graf_gong_carmichael_foss_daltorio_cliffton_2023} used two legs on the same side of a 12-DOF hexapod to grasp a simulated munition buried in the sand. 
However, as more legs are involved in manipulation, fewer are available for locomotion. 
This proposes a unique challenge to walking while grasping, necessitating research on generating and transitioning to gaits with fewer legs. 
In~\cite{whitman2017generating} a hexapod with pincers on its front legs is used for both single-legged and double-legged grasping and transporting of objects. 
They demonstrated fluent gait transitions while simultaneously walking and lifting up a can of WD40 with one leg, a tall box with its front two legs, or a long box below its body with its middle two legs. 
Similarly,~\cite{9982229} proposed a small-scale quadruped robot with leg-mounted manipulators used for both single- and double-legged grasping. 
~\cite{shawkeyframe} presented rapid online gait transitions for a hexapod robot from an alternating tripod gait with six legs, to a quadrupedal walking gait with four hind legs, using the front two legs for manipulation. 
When multiple legs are used together for manipulation, the same end-effector designs that make for predictable contact on the ground (soft, durable, rounded shapes) also ensure good object contact - thus the end-effector does not warrant modification.

However, in order to walk while holding large objects, stance stability adjustments are crucial to prevent excessive pitch of the robots.  
For instance, if a robot picks up a heavy enough object, resulting in a significant shift in its center of mass position, it is important to prevent the robot from toppling over.
This problem was first demonstrated in 1997 when \cite{Melmantis} presented two hexapod prototypes, Melmantis-1 and Melmantis-2, based on their previous work on six-link ``Limb Mechanisms'' to both walk and manipulate~\citep{koyachi1993integrated}.
Later, in 2002, they remotely controlled Melmantis-1 to handle an object using its two front limbs.
There, to maintain balance while raising two legs, the other four legs transformed into a side-by-side symmetric quadruped, expanding the robot's support polygon~\citep{koyachi2002control}.
Similarly,~\cite{booysen2015gait} proposed gait transition methods during grasping for a hexapod.
Besides shifting the robot's CoM and expanding its support polygon while lifting an object, they introduced an automatic pitch correction method while walking.
If the robot was at risk of toppling, it would stop, lean back (raising the front legs and lowering the hind legs), retract its gripper, or even drop the item if the average pitch angle surpassed a given safety limit.
This helped the robot prioritize recovering its own balance first, before grasping the object again and attempting to continue its locomotion.
More recent works have focused on removing the need for hexapods to even stop walking when recovering their balance.
\cite{deng2018object} proposed a method to automatically adjust the robot's gait when lifting heavier objects with two legs or smaller objects with a single leg.
They estimated the object's mass by sensor feedback (embedded in the end-effector) and then calculated joint motions according to the desired CoM trajectory. 
Considering energy efficiency in manipulation,~\cite{ding_yang_2016} optimized the energy used against stability constraints. 

Compared with front-leg manipulation, using two opposing middle legs for grasping can ensure a larger support polygon.~\cite{895266} designed a hexapod with omnidirectional manipulability and mobility. Unlike most hexapods with longitudinally symmetrical left and right legs, their design incorporated six radially symmetrical legs on a hexagonal body, minimizing the overlap between the workspaces of individual legs. Using a set of these radially opposite legs for lifting, stability during walking was greatly improved. 
\cite{whitman2017generating} also observed that the volume of the object that their hexapod could carry using the front/middle two legs was ten times larger than what it could carry with a single leg.
Further, given the modular nature of their hexapod, two of them could be attached together by the body, creating a larger dodecapod capable of using more legs for carrying larger payloads underneath the robot.

This is analyzed with an extension to the single leg case. When the object is larger than the limit of the pincers, the robot can grab the object using two front legs, being fully extended for maximum grasp size, i.e., double-legged grasping (Fig.~\ref{fig:legs}b and Fig.~\ref{fig:grasp_view}b).
If the diameter of the object is smaller than the width i.e., $r\le0.5\,s$, the front legs can always secure the object even with a small $\mu$.
Otherwise, if $r > 0.5\,s$, like the single-leg case it must be ensured that $\mu > tan(\theta)$ where $\theta$ is half the angle subtended by the front legs (Fig.~\ref{fig:grasp_view}b). It follows that: 
\begin{equation}\label{Double_Leg_eq1}
\mu > \frac{r}{l+\frac{s}{2 sin(\theta)}} = \frac{1}{1+\frac{1}{4 sin(\theta)}} \cdot \frac{r}{l}
\end{equation}
where robot width $s$ can be expressed as $s = \frac{l}{2}$ and $sin(\theta)$ can be expressed by $\frac{r}{l}$ based on geometry (shown in Appendix~\ref{appendix1}). It is complex to simplify Eq.~\ref{Double_Leg_eq1}, however, for our values of $\frac{r}{l}$, there is an approximately linear relationship over the relevant range as follows:

\begin{equation}\label{Double_Leg_eq2}
\mu > 0.89 \cdot \frac{r}{l}-0.21
\end{equation}

\subsubsection{Whole-body Grasping}

\begin{figure}[h!]
\begin{center}
\includegraphics[width=\textwidth]{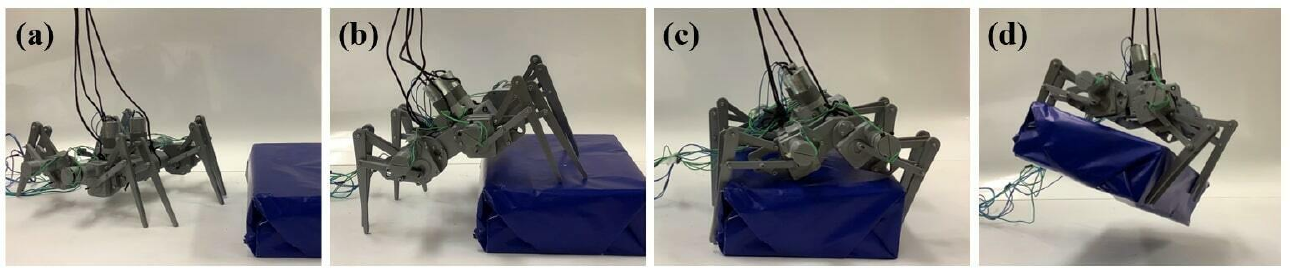}
\end{center}
\caption{Tethered to a boat, crane, or flying platform, the Klann robot presented in~\cite{YifengGong} can (a) walk to the object, (b) climb the object (c) grasp the object, and (d) be pulled by the tether while grasping the object}\label{fig:tether}
\end{figure}

For an even larger object, a quadruped or hexapod robot can sacrifice its locomotion capabilities by using all its legs to achieve whole-body grasping (Fig.~\ref{fig:legs}c and Fig.~\ref{fig:grasp_view}c). 
We observe that for most effective grasping, the robot design must be radially symmetrical. (E.g., the base joints of legs must form a square for a quadruped or a regular hexagon for a hexapod). 
In this case, $\theta$ is dependent on the distance $d$ between the base joints and the robot's centre.
For an n-legged robot, $d$ can be expressed as $d = \frac{s}{2} \cdot sec(\alpha) = \frac{l}{4} \cdot sec(\alpha)$ where $\alpha = \frac{\pi \cdot (n-2)}{2 \cdot n}$.

\begin{equation}\label{Whole_Body_eq1}
\mu > \frac{r}{l+\frac{d}{sin(\theta)}} = \frac{1}{1+\frac{1}{4 sin(\theta)\cdot cos(\alpha)}} \cdot \frac{r}{l}
\end{equation}
where $sin(\theta)$ can be expressed by $\frac{r}{l}$ from geometry (shown in Appendix~\ref{appendix1}). The approximately linear relationship for quadruped whole-body grasping is Eq.~\ref{Whole_Body_eq2} and hexapod whole-body grasping is Eq.~\ref{Whole_Body_eq3}.
\begin{equation}\label{Whole_Body_eq2}
\mu > 0.85 \cdot \frac{r}{l}-0.27
\end{equation}
\begin{equation}\label{Whole_Body_eq3}
\mu > 0.82 \cdot \frac{r}{l}-0.38
\end{equation}

\begin{figure}[htbp]
\begin{center}
\includegraphics[width = 0.8\textwidth]{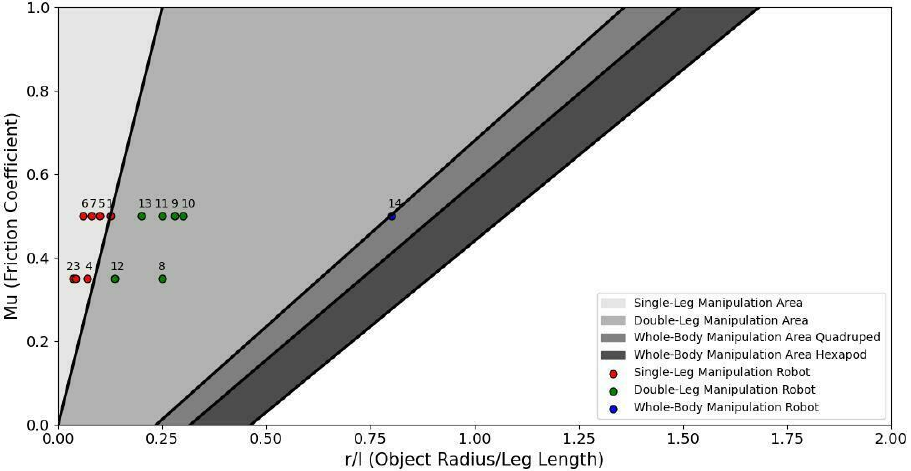}
\end{center}
\caption{The manipulation model analysis depicts the required friction coefficient $\mu$ v.s. object size normalized against robot leg length $r/l$. The data points plotted are listed in Table~\ref{Table of friction coefficients of the manipulators}. 
When not provided in literature, respective friction coefficients are approximated depending on the material of the manipulator surface: For metal $\mu$ is 0.35~\citep{engineeringtoolbox}(static coefficient between steel and polystyrene is 0.3-0.35); and for plastic $\mu$ is 0.5~\citep{engineeringtoolbox}(static coefficient between polystyrene and polystyrene is 0.5)
}\label{fig:GraspAnalysis_Results}
\end{figure}

Our manipulation model analysis is presented in Fig.~\ref{fig:GraspAnalysis_Results}. 
We can infer a strong correlation between our method of analysis and experimental results in real-life examples, helping validate our simplified model and analyses.
Generally speaking, by grasping with a single leg the robot can grasp relatively small objects, depending on the pincer length $l_{p}$.
While double-legged grasping has an extended area for bigger objects than a single leg, most robots can only manipulate small-size objects in the double-legged grasping area since the robots usually cannot fully extend legs for big object manipulation due to the lack of motor torque, joint angles and the position shift of CoM.
In comparison, whole-body grasping can safely grasp objects roughly the robot's size. Hexapods can use all legs to grasp larger objects than quadrupeds if the length of the legs and the width of the robot are the same.

Instead of using some of the legs to grasp an object, the whole robot can behave as a mobile gripper that walks towards an object and grasps it with all legs (i.e., whole-body grasping).
A legged robot will not be able to continue walking with the large object after grasping it with all of its legs, but it can still be transported externally, e.g., via a cable from a boat, crane, or flying platform (Fig.~\ref{fig:tether}).
One of the key advantages there is that, unlike a typical crane end-effector, the robot can walk on the ground to systematically search for the object, and/or reach areas occluded from the crane by overhangs or ledges.
The tether can then be used to recover both the object and robot, after the robot has securely grasped it with all legs.
In this case, the need for walking ends when the object is found, which simplifies the problem of lifting the found object and thus enables the potential for a relatively smaller robot to transport a relatively larger object.
Besides walking and grasping, climbing is another main challenge for whole-body manipulation since the robot must often climb over thean object to then grasp it.
Notably,~\cite{YifengGong} designed a four-DoF walking robot by adding coxa joints on a seven-link Klann mechanism, which is able to climb onto a rectangular box and then grasp it using all legs.
While two (drive) DOFs are responsible for generating periodic gait patterns (Fig.~\ref{fig:tether}a), the other two (lift) DoFs are used to raise the front legs separately and climb onto the object (Fig.~\ref{fig:tether}b) and then securely press its legs inward onto it (Fig.~\ref{fig:tether}c, generating sufficient friction for a secure grasp.
Both the robot and the object can be manually retracted by a tether  (Fig.~\ref{fig:tether}d). 

Given the growing popularity of legged quadrupeds with the team Cerberus from ETH-Zurich winning DARPA's Subterranean Challenge in 2021, as well as the increasing availability of commercial quadruped dog-scale robots like Spot mini, Ghost, ANYmal, and others, manipulation with walking legs is a promising direction.
Even legs may at times be less reliable and precise than industrial manipulators, legged robots can still carry a variety of objects in many relevant scenarios, without requiring an additional expensive manipulator appendages.

\subsection{Dedicated Non-Locomotive Arms}
\label{Dedicated arm}

\begin{figure}
\begin{center}
\includegraphics[width=\textwidth]{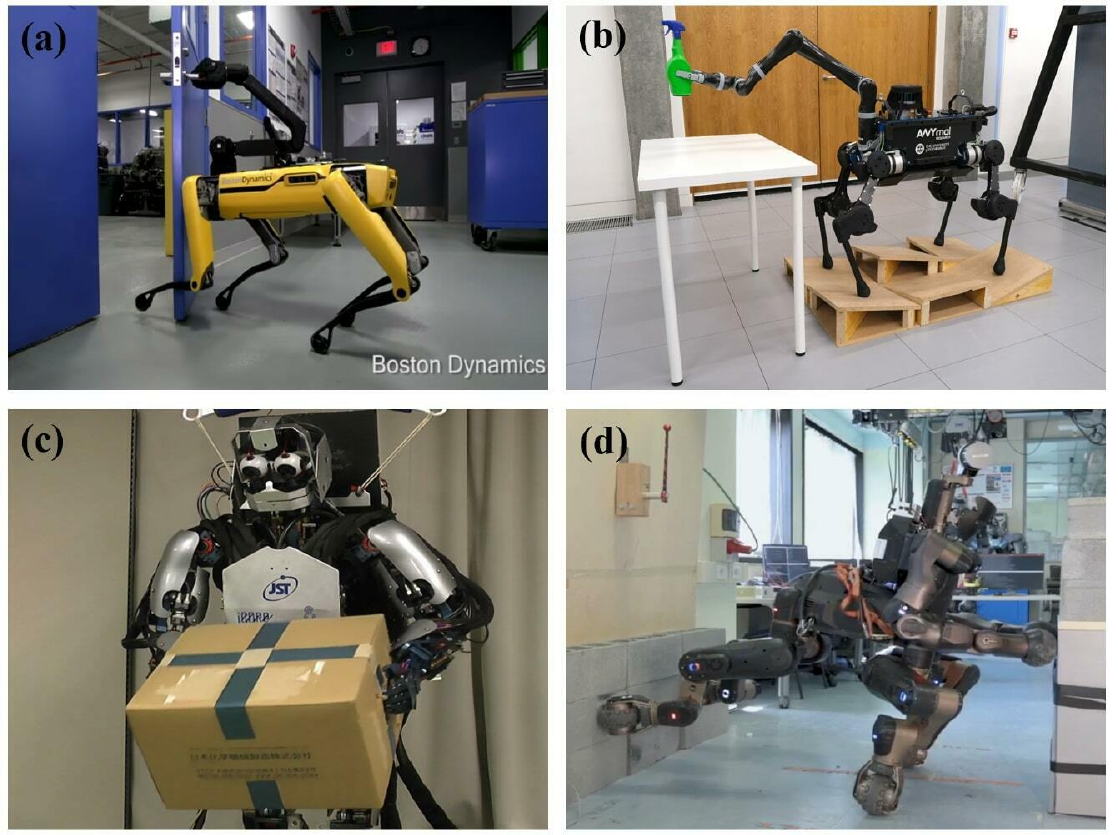}
\end{center}
\caption{Legged robots equipped with dedicated non-locomotive arms can perform various useful tasks, such as (a) opening doors (Spot~\cite{spot2018door}), (b) manipulation on structured terrain (ANYmal~\cite{ferrolho2020optimizing}), and (c) lifting (CB-i~\cite{gams2015accelerating}) or (d) pushing large objects (CENTAURO~\cite{polverini2020multi})}\label{fig:DedicatedArmPaperImages}
\end{figure}

More sophisticated tasks may require dexterous end-effectors that are more delicate and multi-functional than legs.
In this section, we discuss systems in which a dedicated arm was added to a legged robot that is different from the walking legs.
Because they are not used for locomotion, these dedicated appendages can be specialized for fine sensing, grasping, and manipulation, or can be designed to lift heavier objects.
However, equipping a legged robot with a dedicated arm will also present some challenges (e.g., stability, and cooperation between arms and legs) as discussed in the following paragraphs.

Here, we will reserve the term ``legs'' for appendages that are meant to carry a significant portion of the body's weight during locomotion, and non-locomoting appendages as ``arms''.
Sometimes these robots have a giraffe-like body plan with a sensorized arm on top for high reach, however, they also include bipeds~\citep{maniatopoulos2016reactive}, centaurs~\citep{schwarz2017nimbro}, octopuses~\citep{mazzolai2012soft} and many different form factors.
Note that any brachiating robots~\citep{lo2017model} (robots that swing like apes) would be a special case here, but to our knowledge at this time, few, if any brachiating robots have been demonstrated for manipulation of the environment and could rather be categorized as adding an inverted gripper for the ground to the legs.

Thus, arguably the most impressive robotic tasks performed by legged robots have used an added arm.
Bigdog was first demonstrated to be capable of throwing a 16.5 kg~block over 17 feet using a hydraulic manipulator, including the gripper, installed on top of its body~\citep{murphy2012high}.
In 2016, Boston Dynamics equipped the Spotmini with an electric robotic arm.
With this, Spotmini could complete everyday tasks like opening doors or stabilizing an object in the world frame while the robot's base moved around, showing great cooperation between the legs, body, and arm~\citep{spot2018door} (Fig.~\ref{fig:DedicatedArmPaperImages}a). 

Using a legged platform for arms enables adaptations that are not possible with wheeled robots.
A legged robot with an arm can dynamically manipulate objects on challenging terrain, including stairs and ladders.
For example,~\cite{ferrolho2020optimizing} developed a control framework allowing ANYmal-B to complete pick-and-place tasks on various rough terrains (Fig.~\ref{fig:DedicatedArmPaperImages}b).
\cite{ ma2022combining} later combined a learning-based locomotion policy with a model-based manipulation controller to make ANYmal-C track a desired end-effector position for its arm when trotting on challenging terrains.
Moreover, compared to wheeled platforms, legged platforms can significantly improve the arm's manipulation capabilities in terms of reachability (workspace), payload capacity, and manipulation speed since legged platforms often have more degrees of freedom resulting in finer control of the base's position (especially height) and orientation.
Specifically, as stated in~\cite{murphy2012high}, the knees and hind legs can assist the arm during lifting and throwing motions.
Finally, legged platforms are able to change foot positions for bracing, as discussed in Section~\ref{Pushing}.

Another advantage is that giraffe-like platforms can achieve baseline performance with almost independent controllers for locomotion and manipulation, and then advance to more whole body optimization as needed.
This is different from standard mobile manipulation with a wheeled platform, where the base platform cannot reconfigure as much during manipulation.
\cite{zimmermann2021go} combine Boston Dynamics' Spotmini with a light-weight robotic arm and a gripper. 
They propose a straightforward trajectory optimization method based on a simple dynamic model of the platform, where Spotmini is treated as a black box.
Experiments demonstrate that the combined platform is capable of performing dynamic grasping tasks using a feedforward controller.
Contrary to model-based learning methods, model-free learning methods do not incur an increase in computational cost due to the complexity of the underlying dynamic model.
\cite{ma2022combining} combine a learning-based policy for locomotion with a model-based controller for manipulation.
They train the locomotion policy with known random wrench disturbances, obtained from the dynamic plans of the model-based manipulator controller.
Without retraining, the combined controller adapts to a variety of manipulator configurations.
In contrast,~\cite{yao2022transferable} use a policy based on reinforcement learning to discover the dynamic representation of different arms for rapid migration, and then feed the estimated dynamic parameters to the low-level model-based locomotion controller.
In addition to hierarchical control pipeline for legged manipulators,~\cite{fu2022deep} train a unified policy for the whole system with reinforcement learning. 
Their policy demonstrates coordination between the legs and arms as well as dynamic control over them.

Furthermore, an increasing number of works have considered giraffe-like platform as a whole-body optimization problem.
Relying on online hierarchical optimization, ANYmal-B can perform reactive human-robot collaboration and body posture optimization to increase the manipulability of its onboard arm~\citep{bellicoso2019alma}.
More recently,~\cite{ewen2021generating} improved their work from immutable contact force to online optimized force trajectory to ensure dynamic feasibility and stability of the platform while~\cite{sleiman2021unified} presented the first holistic MPC framework combining dynamic locomotion and manipulation for ANYmal-C.
\cite{chiu2022collision} extended the whole-body MPC controller to generate collision-free motions during the coordination of locomotion and manipulation.
\cite{ morlando2022nonprehensile} proposed a whole body controller to transport a non-prehensile object with a giraffe-like platform.

An onboard arm can be specialized to reach a large volume around the robot, which the legs could not easily reach without drastic morphological changes to the robot.
In other words, if the arm is mounted on the top of the body and can rotate fully around a vertical axis, its actual workspace can approach a sphere and
completely encompass the body of the robot.
Therefore, a specialized arm can be used to tackle complex real-world tasks such as rotating a handwheel, handling articulated objects, and unlatching a door.~\citep{ferrolho2022roloma, mittal2022articulated, sleiman2021unified}, which would be much harder to approach using locomotive legs.
In addition, an onboard arm can reach the legs themselves, which may prove useful if the legs need to be repaired or untangled.

However, because the legs and body occlude the arm's workspace, the arm's footprint on the ground may be smaller than that of the combined legs.
Thus, if the goal is primarily to manipulate objects on the floor, legs may be better suited~\citep{heppner2015laurope}.
Furthermore, the statically stable workspace of the arm may be a smaller subset of the kinematic workspace.
In other words, if a long arm is heavy or holding a heavy object, the configuration of the legs will be crucial in preventing the robot from tipping over.
For example, the HyQ robot is equipped with a hydraulic manipulator (HyArm~\citep{rehman2016towards}) at the front-middle of the body.
The HyArm weighs 12.5 kg and is capable of carrying up to 10 kg of payload.
The arm placement dramatically shifts the robot's CoM and affects its balance.
\cite{rehman2016towards} proposed a new control framework through optimizing the Ground Reaction Forces (GRFs) of a quadruped and compensating for external/internal disturbances induced by the manipulator.
The integrated payload estimator module is utilized to estimate an unknown payload carried by the arm, in order to update the robot's center of mass (CoM) position and control its stability.

Towards even more human-like behavior, some robots, specifically bipeds and centaurs, have two onboard arms.
Different from the double-legged grasping discussed in Section~\ref{Multi-Legged-Manipulation}, dedicated dual-arm setups have more degrees of freedom and more flexible end effectors.
Dual arms have the advantage of being able to directly mimic human motions, either repetitive motions or human tele-operated motions~\citep{kulic2012incremental, zuher2012recognition, gams2022manipulation, kim1999development,zhou2022teleman}.
In addition, dual arms can work together for large object manipulation~\cite{gams2015accelerating} (Fig.~\ref{fig:DedicatedArmPaperImages}c) and hand-to-hand manipulation~\cite{vahrenkamp2009humanoid}.
However, as a result of their hefty upper body, centaurs may experience instability when manipulating larger objects.
\cite{ polverini2020multi} presents a control architecture for pushing a heavy object by optimizing the contact force with the environment (Fig.~\ref{fig:DedicatedArmPaperImages}d).
For additional information on manipulation with bipeds and centaurs, we refer the reader to~\cite{gams2022manipulation}, which studies humanoid robot learning for manipulation capabilities, and~\cite{rehman2018centaur}, which provides an overview of centaur-style robots with dual arm manipulator systems.
Another way of tackling the stability problem for very large objects is to have multiple robots working together, as described in the next section.

\subsection{Legged Teams for Manipulation}
\label{Legged team}

\begin{figure}[h]
    \centering
    \includegraphics[width = 1\linewidth]{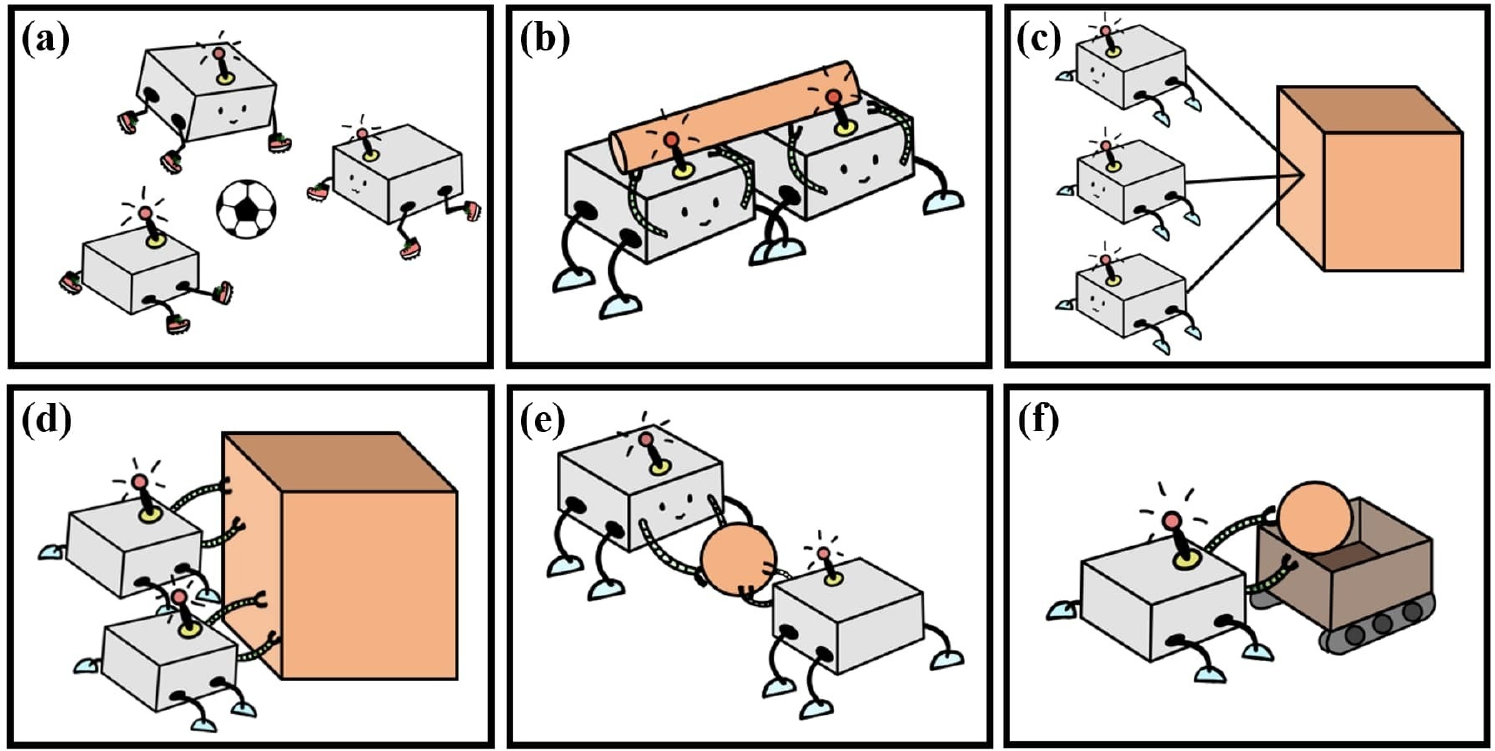}
    \caption{Main classes of multi-Robot manipulation works to date: (a) soccer playing~\citep{hu2001reactive}, (b) carrying~\citep{hara1999motion}, (c) towing~\citep{christensen2016let}, (d) pushing~\citep{mataric1995cooperative}, (e) lifting~\citep{fujita1998tight}, and (f) heterogeneous robot teams~\citep{govindaraj2021building}. Image by and courtesy of Yixian Qiu.}
    \label{fig:arm}
\end{figure}

\begin{figure}[htbp]
\begin{center}
\includegraphics[width=\textwidth]{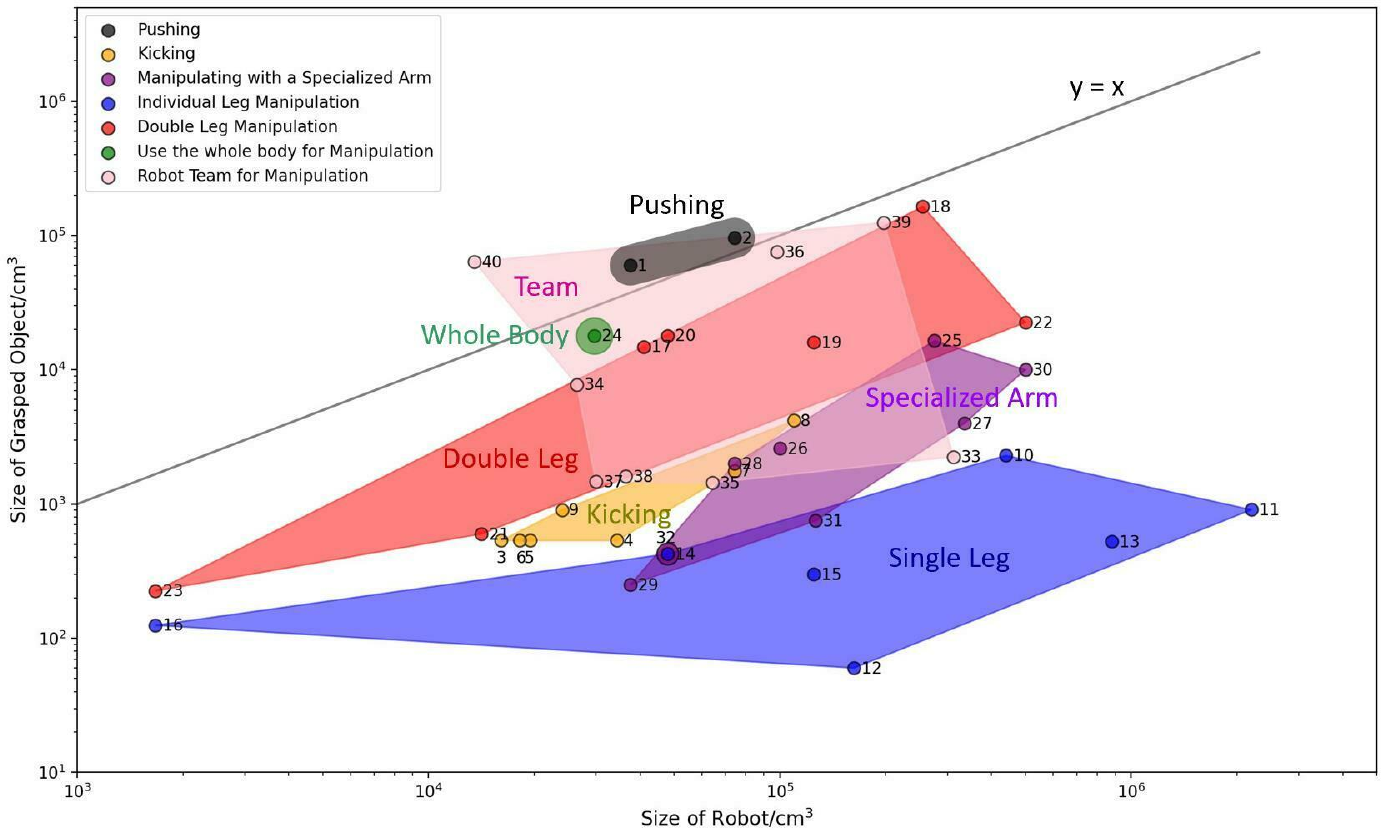}
\end{center}
\caption{Carrying capacity of different manipulation methods (maximum object size v.s. maximum robot size). The data points plotted are listed in Table~\ref{Table of the maximum size of the robots and maximum carrying capacity}. If the point is above the reference line (object size is equal to robot size), the robot can manipulate an object larger than itself. 
This data was obtained through literature sources. 
For cases when this wasn't explicitly provided, we approximated the object size from the published figures or approximated their volume using the density of water.
Robot size is determined by the maximum dimensions.}\label{fig:GraspingCapacity}
\end{figure}

A team of legged robots may work together to manipulate objects that are too large, heavy, or awkwardly shaped for a single robot.
Although the work discussed in previous sections has shown that individual robots can manipulate a wide variety of objects in diverse applications, there are some key restrictions in their abilities due to gripper size, ground support based on body/leg dimensions, and actuator torque limits (Fig.~\ref{fig:GraspingCapacity}).
On the other hand, a team of robots can bring additional support/grasping points to securely hold a large object, as well as distribute the object weight among team members to increase the total manageable payload capacity and/or size~\citep{hara1999motion,hamed2019hierarchical,chang2022design}.
Furthermore, even in cases where a single robot is capable of manipulating an object, the object can be cooperatively relayed from one robot to the next, thereby expanding the overall range and speed of the operation.
For example,~\cite{hu2001reactive} leveraged this principle by using robots to pass a football (Fig.~\ref{fig:arm}a).
More generally, collaboration is especially useful in cases where multiple general purpose robots are available rather than a large, special-purpose robot.  

One way that legged robots can collaboratively manipulate a large and/or heavy object is by lifting and then carrying it together.
The team of robots must first move to designated locations around the object, after which they can perform coordinated motions for lifting. 
~\cite{chang2022design} proposed to detect the object using a vision system that recognizes specific markers on the object to determine its location, after which a team of  legged robots can move towards it until they reach appropriate collaborative lifting positions.
In this case, the robots used a sideways gait to move, which was synchronized using a fuzzy motion control algorithm that selects the appropriate repeat cycles for the fast and slow pre-planned gaits to cooperatively transport the object toward the destination.
This is similar to how humans adjust their posture to avoid injury when lifting/carrying objects together.
It is essential for lifting robots to receive feedback about the orientation of the object, which ensures that they do not apply any unwanted redundant forces onto it.
To this end, ~\cite{hara1999motion} implemented coordinated lifting while introducing a leader-follower system using two TITAN VIII quadruped robots (Fig.~\ref{fig:arm}b).
There, the leader was tasked with maintaining a course to a given destination, while the follower(s) adjusted their own path based on feedback of the object's relative position without explicit communications with the leader.
Instead of correcting the follower's positions based on the state of the object,~\cite{inoue2003cooperative} proposed another approach on a small team of HOAP-1 humanoid robots, where only the follower robot(s) corrected their position relative to the leader, which still mainly maintained a path towards the goal.
~\cite{hawley2019control} used NAO humanoid robots to demonstrate robot-object-robot constraints (i.e., relative safe distances between robots and object) during cooperative manipulation using a Zero Moment Point controller that maintained the dynamic stability of the system.
Once the object was lifted, the entire system (robots + object) could work together as a large multi-legged robot, with the object behaving as its body.
\cite{mcgill2011cooperative} demonstrated a robotic system composed of two DARwIn-OP humanoids tasked with manipulating and transporting a stretcher system.
After picking up the stretcher, the robots walked together as a ``virtual quadruped'' with careful synchronization between their legs.
This synchronization relied on a two-stage motion controller, which consisted of a step controller and a walking controller to harmonize the movement of the robots.
In particular, the walking controller guided the legs' lifting motions by generating foot and torso trajectories and maintaining the robots' and the object's stability.

As objects get larger, it can become difficult for robots to lift or carry them on their backs: instead, an easier approach relies on dragging an object along the ground (assuming it is flat enough to do so, and the object can be dragged safely).
A team of legged robots can tow such objects by attaching them to their bodies using wires, cables, or rigid rods, similar to how horses pull a wagon or mules a plow.
\cite{boston2019mush} famously demonstrated how 10 Spot robots can tow a truck on a shallow uphill slope (approximately 1 degree of inclination).
In general, the towing performance and efficiency of a legged robot team depend on a variety of factors such as team size, locomotion gait, and surface interactions.
For example, a walking gait generates a relatively constant pulling force as compared to a running gait. 
That is, walking is a smooth gait in which at least a few legs are always in contact with and pushing/pulling against the ground, while running is an impulsive gait that generates pulling forces in bursts as the legs hit the ground.
~\cite{christensen2016let} studied and compared the towing performance of walking and running teams of hexapods (Fig.~\ref{fig:arm}c).
Overall, they showed that walking robots have a better load sharing ability than running robots.
As the number of robots increases, their results show that walking robots also outperform running ones in terms of peak pulling force.
Similar to robot-ground interactions, robot-robot interactions also affect towing performance.
The physical interactions between robots, which depend on the type of object connection used, can be classified into hard (e.g., rigid rods), soft (e.g., deformable objects), and hybrid (e.g., wires, cables), and naturally affect the complexity of the underlying system dynamics and the control problem.
The object can be connected to the robots at single or multiple attachment points, where more points offer the system greater controllability during manipulation.
Unlike rigid rods, cables can allow the system to change its effective dimensions (footprint).
For instance, if the team has to transport a cable-towed object from one place to another in a cluttered environment, it may need to avoid obstacles and travel through narrow spaces.
Due to their agile nature, legged robots can quickly reorient and reposition themselves to change the effective footprint of the overall system by letting the cables switch between taut and slack (e.g., approach the object when needed to reduce the effective footprint of the overall convoy).
Using this feature of cables,~\cite{Yang2022CollaborativeNA} demonstrated their framework for the navigation and collaborative manipulation of a cable-towed load using a team of Mini Cheetah and two A1s through narrow, cluttered, and diagonal spaces.

Communication between legged robots is another crucial aspect during multi-robot cooperative manipulation.
If robots cannot exchange information, each robot can only ever rely on partial knowledge of its local environment (and, more importantly, of other robots' states and strategies/intents).
This often results in robots making independent/uncoordinated decisions, leading to worse collaboration.
\cite{mataric1995cooperative} evaluated the performance of a team of two hexapod robots in a box-pushing task with and without cooperative communication (Fig.~\ref{fig:arm}d). 
Experiments demonstrated that the team cannot complete the task at all in the absence of communications.
In contrast, by communicating with each other (here, sharing sensor data), transient faults of one of the robots can be largely compensated for by the other, building a form of resilience in the system.
However, communications among robots are typically expensive in practice, and limitations usually arise on bandwidth and range in practice.
In order to improve communication efficiency, it is critical to share only the most pertinent information, and to limit communication to times where new/update information needs to be propagated among agents.
For example,~\cite{fujita1998tight} proposed a framework for event-driven control that can initiate communication only when the task enters a problematic state, significantly improving the execution time of a two-hexapod box-lifting task (Fig.~\ref{fig:arm}e). 
In addition, cooperative communication can allow robots to select and enact distinct synergistic roles, and thus contribute to a specific task to the best of their (potentially heterogeneous) abilities.
\cite{hu2001reactive} proposed an approach that relies on explicit and implicit communication allowing a team of Sony AIBO robots to play different roles in a football competition, including kicking, dribbling, passing the ball, intercepting, and scoring.
There, in order to optimize the performance of the team, robots are able to dynamically swap roles during the game based on the situation at hand.

Finally, teams of heterogeneous robots can take advantage of different of synergistic motion or manipulation capabilities (e.g., legged, wheeled, and flying robots, with dedicated arms or using their legs/body for manipulation).
Benefiting from complementary capabilities of individual robots, system efficiency and robustness can be increased by autonomy, parallelization, and functional redundancy. 
In particular, heterogeneous teams that include legged robots have been proposed for different aspects of planetary exploration missions.
\cite{govindaraj2021building} presented a multi-robot system consisting of a rover and a hexapod robot with a movable gantry for load manipulation and 3D printing towards the construction of lunar infrastructure (Fig.~\ref{fig:arm}f).
They demonstrate the system's cooperative manipulation and transport capabilities by transporting two objects to different locations simultaneously using Mantis and Veles robots.
In addition, and combining the specific stability and speed advantages of wheels and legs over different types of terrains,~\cite{cordes2011lunares,cordes2012rimres} proposed the use of a wheeled rover to transport a legged robot for enhanced extra-terrestrial mobility in craters exploration.
Then, using specialized grippers on its front legs, the legged robot can climb down the crater to collect geological samples.
\cite{kiener2007cooperation} also proposed a method that enables a humanoid robot mounted on a wheeled robot to track a soccer ball and kick it into a goal.
The entire process is broken down into subtasks that are assigned to the individual robots by underlying finite state machines.
A similar approach was used to command a collection of humanoid robots to clean a table by picking up cans and dropping them into a trash box in~\cite{lim2008multiple}.

\section{Open Avenues for Research}

\subsection{Specific Challenges in Legged Robot Manipulation}
The current progress of legged robot manipulation poses many theoretical and experimental challenges to be overcome to make robots more efficient and effective. In this section, we highlight and summarize three key challenges that are common to multiple types of legged robot manipulation in the open avenue: utilizing multi-functional appendages, exploiting challenging terrain, and learning-based control for legged manipulation. By providing a detailed discussion of these challenges, we believe that these challenges can be interesting opportunities for new research.

\subsubsection*{\textbf{Utilizing Multi-functional Appendages}}

A robot that can easily adapt its appendages for use as both legs and arms can make better use of all available degrees of freedom in any unpredictable field scenario. 
However, finding the best body configuration for a particular application remains an open question, where we cannot solely rely on biological inspiration. 
For example, while many biological quadrupeds have specialized front limbs for a variety of tasks (swimming, digging, grasping etc.), adding a fifth limb seems a more difficult task for evolution than it does for robot designers. 
A variety of possibilities can result from such an adaptation, including improved stability and strength, recovering from limb failure, and better dexterity.
Therefore, a comprehensive understanding of how weight, power consumption, and redundancy affect multi-functionality in robots can help inform such designs.

New design principles and strategies will be key in developing such multi-functional limbs. 
As legs, they must be robust enough to support a heavy body and resist repeated impacts with hard substrates during locomotion. 
Since legged robots are often envisioned to replace wheeled robots in unstructured outdoor environments, their legs should also survive dirt and water ingress. 
As arms or manipulators, they must be capable of dexterous manipulation. 
To achieve this, haptic feedback and/or refined dexterous grippers can be employed. Additionally, this also aids in locomotion as haptic sensing provides information as to the robot's interaction with the environment (e.g. ground reaction force). 
However, operating such delicate sensors and grippers in these load-bearing applications remains an open challenge. 
While current research on walking legs still uses simple grippers without sensor feedback (as discussed in Section~\ref{Manipulation With Walking Legs}), robust multi-functional appendages will most likely pave the way to a myriad of exciting research and applications, and will create new opportunities for materials research, mechanical design, as well as for advanced data filtering and processing.

\subsubsection*{\textbf{ 
Exploiting Challenging Terrain
}}
Legged robots can be especially valuable in hazardous or unstable environments that are unsuitable for humans, such as search and rescue tasks in disaster relief scenarios, remote sample collection, or habitat construction for space exploration missions.
While legged robots exhibit impressive mobility on unstructured and/or inclined terrain, even up to vertical climbing~\citep{hong2022agile}, manipulation in these environments requires special considerations. 
If the terrain is inclined, soft or granular, ground reaction forces may cause slippage or sinkage during critical manipulation tasks. Sub-surface support may be insufficient, causing the robot to trip or stumble.
While some works have focused on mitigating these perturbations during locomotion in a reactive manner~\citep{sartoretti2018central}, the effect of such challenging terrain on manipulation abilities has been considered less often to date.

Terrain models will likely be important. 
While stabilizing the manipulator using reactive control may improve performance by incorporating appropriate terrain/interaction models into a control framework (e.g., Model Predictive Control), a robot could predict how the terrain will be modified during the manipulation. This will enable the legs to choose configurations, footholds, and timing to maximize success. 
Planning further ahead, it may be possible to circumvent any issues in advance, leading to improved and more reliable performances during challenging manipulation tasks.

\subsubsection*{\textbf{Learning-Based Control for Legged Manipulation}}

Even for pure locomotion, model-based control methods for online planning are burdened by expensive computational costs from the high number of degrees of freedom.
To learn more about the control method involved in model-based approaches for legged manipulation, we suggest referring to the survey~\cite{chai2022survey} for an up-to-date overview.
To avoid high computational costs, one of the solutions that the locomotion community has started investigating is learning-based control.
In learning-based approaches, heavy computation is mostly focused upon within the training phase, yielding (often time-invariant) policies that can be used at high frequencies.
However, the robustness and generalizability of these policies are often lower, as many learning-based methods rely on neural networks for function approximation, which famously obfuscate what was truly learned by the system behind a ``black box.''
In the long-run, these issues may be mitigated by some clever training techniques, such as privileged or imitation learning~\citep{chen2020learning} or by provably-robust learning approaches~\citep{choromanski2020provably}.
In the shorter-term, many approaches have been proposed to embed learned policies within a robust high-level control framework to ensure the system's safety at run-time, with minimal effect on the overall performance and time complexity.

Learning-based methods have been used effectively and have demonstrated promising results for pure locomotion tasks~\citep{miki2022learning}, but few approaches, if any, have looked at legged manipulation tasks.
We believe that learning a high-dimensional controller for whole-body manipulation is a key long-term objective in this area since current model-based controllers (e.g., MPC) still struggle with this task.
In the longer-term, future research will also investigate an end-to-end control pipeline for the entire system, where the input to the neural network is raw sensor data (e.g., from cameras/LiDAR, proprioceptive, inertial, and haptic sensors), and the output is the final control output (e.g., joint angles or desired torques).
Even though such an approach will inevitably further reduce explainability, such all-in-one control pipelines are much easier to implement on real robots and most likely represent the next frontier for high-performance, versatile legged robot controllers for advanced legged manipulation tasks.

\subsection{New Applications for Legged Robot Manipulation}

The combination of legged locomotion and the ability to manipulate objects will also open up the possibility for new robotic deployments for manipulation tasks in both industry and daily life.

\subsubsection*{\textbf{Challenging Environment Applications}}

Combined legged locomotion and manipulation literally and figuratively open new doors for robotic access in challenging environments. 
While wheeled robots may be sufficient for many tasks where terrains can be initially constructed or modified for wheels (e.g., offices or warehouses), many applications remain out-of-reach of wheeled platforms (and might never be the solution).
On the other hand, legged robots can enable manipulation tasks in a wide array of environments and conditions.
For example, instead of sending humans into disaster sites after extreme events such as nuclear disasters, earthquakes, or floods, legged robots can be deployed to explore hazardous areas and obtain early situation reports~\citep{siegwart2015legged}.
Legged manipulation robots can further help by manipulating the environment to sanitize or secure areas before humans can enter, through teleoperation, or eventually, through autonomous operation.

Conversely, manipulation can help increase the reachability of legged locomotion, by allowing the robot to modify its surrounding to clear obstacles~\citep{lu2020autonomous}.
A common impediment to the increasing use of mobile robots is that if their pre-defined path is blocked, the only option often is to find an alternative trajectory, which does not always exist, making some areas unreachable.
Legged manipulation robots can move away obstacles or modify their surrounding (e.g., move soil to create a ramp, or to reduce the steepness of an incline) to clear/create a path, showing increased potential for search-and-rescue, domestic cleaning, and elder care applications.
In the long run, the fundamental advantage of legged robot manipulation will be in allowing robots to autonomously understand their environment to navigate and manipulate their surroundings, for the purpose of completing a high-level objective with minimal human supervision/action.

\subsubsection*{\textbf{Home Robotic Applications}}

In certain industrial settings (e.g., warehouses or shop floors), the environment may be simple and flat, or could even be specifically designed for robots.
In contrast, home environments are quite complex and constantly changing (e.g., stairs, gardens, and grounds with many obstacles or opportunities for small accidents), for which legged robots are naturally better suited.
We believe that home robotics will be one of the most important future applications for legged robot manipulation~\citep{linghan2022dogget} because they will be able to improve living quality, safety, and emergency response.
For example, disabled and elderly people, who have difficulty performing basic daily activities independently, may specifically benefit from a cheaper, on demand legged robotic home assistant.
Legged robots can help with laundry and carrying heavy payloads (e.g., moving furniture), and can even help with other everyday tasks such as preparing food and managing medications for people who suffer from cognitive disorders.
Finally, a legged robot may even help a person after a home accident, such as an elder person who has fallen down or someone experiencing a stroke, by calling for help from a phone, bringing medical supplies, helping them to stand back up or to reach a safe area in the house, or even by directly performing CPR!

\color{black}
\section{Conclusion}

This review provides a broad overview of recent work in legged robot manipulation, which can be classified into four main categories: non-prehensile manipulation, manipulation with walking legs, manipulation with a specialized arm, and legged team manipulation.
Each of these approaches results in different robot designs and has different advantages.
If the object to be moved is small or light, a dedicated non-locomotive arm or leg-mounted gripper can be easily added to an existing robot.
Larger objects can be moved if multiple legs or multiple robots work together.
Dexterity is also important and requires both precise actuation and sensing.
Another consideration is specificity vs generality, in which some end effectors are useful only for a small range of objects, while other approaches such as pushing can be implemented on any robot.
With the manipulation methods mentioned in the review, legged robots are expected to work on various applications in tough terrains by actively interacting with their surroundings instead of passive interaction only.
We believe legged manipulation is one of the next frontiers of robotic research.
Our hope is that this review provides helpful examples and motivates additional research effort in all these exciting areas.

\section*{Acknowledgements}

This work was partly supported by the Singapore Ministry of Education Academic Research Fund Tier 1.
This work was partly supported by the Office of Naval Research under a Young Investigator Award under Grant N000141912138 and in part by the Strategic Environmental Research and Development Program (SERDP) SEED under Grant MR19-1369. (Corresponding author: Kathryn Daltorio.)

The authors would like to thank Swasti Khurana and Carla Macklin for helping proofread the entire manuscript.

\bibliographystyle{Frontiers-Harvard} 
\bibliography{references.bib} 

\begin{thebibliography}{135}
\providecommand{\natexlab}[1]{#1}
\expandafter\ifx\csname urlstyle\endcsname\relax
  \providecommand{\doi}[1]{doi:\discretionary{}{}{}#1}\else
  \providecommand{\doi}{doi:\discretionary{}{}{}\begingroup
  \urlstyle{rm}\Url}\fi
\providecommand{\selectlanguage}[1]{\relax}
\providecommand{\bibAnnoteFile}[1]{%
  \IfFileExists{#1}{\begin{quotation}\noindent\textsc{Key:} #1\\
  \textsc{Annotation:}\ \input{#1}\end{quotation}}{}}
\providecommand{\bibAnnote}[2]{%
  \begin{quotation}\noindent\textsc{Key:} #1\\
  \textsc{Annotation:}\ #2\end{quotation}}

\bibitem[{Acosta-Calderon et~al.(2008)Acosta-Calderon, Mohan, Zhou, Hu, Yue,
  and Hu}]{acosta2008modular}
Acosta-Calderon, C.~A., Mohan, R.~E., Zhou, C., Hu, L., Yue, P.~K., and Hu, H.
  (2008).
\newblock A modular architecture for humanoid soccer robots with distributed
  behavior control.
\newblock \emph{International Journal of Humanoid Robotics} 5, 397--416
\bibAnnoteFile{acosta2008modular}

\bibitem[{Altuzarra et~al.(2011)Altuzarra, {\c{S}}andru, Pinto, and
  Petuya}]{altuzarra2011symmetric}
Altuzarra, O., {\c{S}}andru, B., Pinto, C., and Petuya, V. (2011).
\newblock A symmetric parallel sch{\"o}nflies-motion manipulator for
  pick-and-place operations.
\newblock \emph{Robotica} 29, 853--862
\bibAnnoteFile{altuzarra2011symmetric}

\bibitem[{Aoi et~al.(2017)Aoi, Manoonpong, Ambe, Matsuno, and
  W{\"o}rg{\"o}tter}]{aoi2017adaptive}
Aoi, S., Manoonpong, P., Ambe, Y., Matsuno, F., and W{\"o}rg{\"o}tter, F.
  (2017).
\newblock Adaptive control strategies for interlimb coordination in legged
  robots: a review.
\newblock \emph{Frontiers in neurorobotics} 11, 39
\bibAnnoteFile{aoi2017adaptive}

\bibitem[{Behnke and St{\"u}ckler(2008)}]{behnke2008hierarchical}
Behnke, S. and St{\"u}ckler, J. (2008).
\newblock Hierarchical reactive control for humanoid soccer robots.
\newblock \emph{International Journal of Humanoid Robotics} 5, 375--396
\bibAnnoteFile{behnke2008hierarchical}

\bibitem[{Bellicoso et~al.(2018)Bellicoso, Bjelonic, Wellhausen, Holtmann,
  G{\"u}nther, Tranzatto et~al.}]{bellicoso2018advances}
Bellicoso, C.~D., Bjelonic, M., Wellhausen, L., Holtmann, K., G{\"u}nther, F.,
  Tranzatto, M., et~al. (2018).
\newblock Advances in real-world applications for legged robots.
\newblock \emph{Journal of Field Robotics} 35, 1311--1326
\bibAnnoteFile{bellicoso2018advances}

\bibitem[{Bellicoso et~al.(2019)Bellicoso, Kr{\"a}mer, St{\"a}uble, Sako,
  Jenelten, Bjelonic et~al.}]{bellicoso2019alma}
Bellicoso, C.~D., Kr{\"a}mer, K., St{\"a}uble, M., Sako, D., Jenelten, F.,
  Bjelonic, M., et~al. (2019).
\newblock Alma-articulated locomotion and manipulation for a
  torque-controllable robot.
\newblock In \emph{2019 International Conference on Robotics and Automation
  (ICRA)} (IEEE), 8477--8483
\bibAnnoteFile{bellicoso2019alma}

\bibitem[{Bertoncelli et~al.(2020)Bertoncelli, Ruggiero, and
  Sabattini}]{bertoncelli2020linear}
Bertoncelli, F., Ruggiero, F., and Sabattini, L. (2020).
\newblock Linear time-varying mpc for nonprehensile object manipulation with a
  nonholonomic mobile robot.
\newblock In \emph{2020 IEEE International Conference on Robotics and
  Automation (ICRA)} (IEEE), 11032--11038
\bibAnnoteFile{bertoncelli2020linear}

\bibitem[{Billard and Kragic(2019)}]{billard2019trends}
Billard, A. and Kragic, D. (2019).
\newblock Trends and challenges in robot manipulation.
\newblock \emph{Science} 364, eaat8414
\bibAnnoteFile{billard2019trends}

\bibitem[{Booysen and Reiner(2015)}]{booysen2015gait}
Booysen, T. and Reiner, F. (2015).
\newblock Gait adaptation of a six legged walker to enable gripping.
\newblock In \emph{2015 Pattern Recognition Association of South Africa and
  Robotics and Mechatronics International Conference (PRASA-RobMech)} (IEEE),
  195--200
\bibAnnoteFile{booysen2015gait}

\bibitem[{BostonDynamics(2018)}]{spot2018door}
[Dataset] BostonDynamics (2018).
\newblock Hey buddy, can you give me a hand?
\newblock \url{https://www.youtube.com/watch?v=fUyU3lKzoio}
\bibAnnoteFile{spot2018door}

\bibitem[{BostonDynamics(2019)}]{boston2019mush}
[Dataset] BostonDynamics (2019).
\newblock Mush, spot, mush!
\newblock \url{https://www.youtube.com/watch?v=OnWolLQSZic}
\bibAnnoteFile{boston2019mush}

\bibitem[{Chai et~al.(2022)Chai, Li, Song, Zhang, Zhang, Liu
  et~al.}]{chai2022survey}
Chai, H., Li, Y., Song, R., Zhang, G., Zhang, Q., Liu, S., et~al. (2022).
\newblock A survey of the development of quadruped robots: Joint configuration,
  dynamic locomotion control method and mobile manipulation approach.
\newblock \emph{Biomimetic Intelligence and Robotics} 2, 100029
\bibAnnoteFile{chai2022survey}

\bibitem[{Chang and Tao(2022)}]{chang2022design}
Chang, C.-W. and Tao, C.-W. (2022).
\newblock The design and the development of a biped robot cooperation system.
\newblock \emph{Processes} 10, 1350
\bibAnnoteFile{chang2022design}

\bibitem[{Chebotar et~al.(2017)Chebotar, Hausman, Zhang, Sukhatme, Schaal, and
  Levine}]{chebotar2017combining}
Chebotar, Y., Hausman, K., Zhang, M., Sukhatme, G., Schaal, S., and Levine, S.
  (2017).
\newblock Combining model-based and model-free updates for trajectory-centric
  reinforcement learning.
\newblock In \emph{International conference on machine learning} (PMLR),
  703--711
\bibAnnoteFile{chebotar2017combining}

\bibitem[{Chen et~al.(2020)Chen, Zhou, Koltun, and
  Kr{\"a}henb{\"u}hl}]{chen2020learning}
Chen, D., Zhou, B., Koltun, V., and Kr{\"a}henb{\"u}hl, P. (2020).
\newblock Learning by cheating.
\newblock In \emph{Conference on Robot Learning} (PMLR), 66--75
\bibAnnoteFile{chen2020learning}

\bibitem[{Chen et~al.(2018)Chen, Kong, Li, Wang, and Lin}]{POE}
Chen, G., Kong, L., Li, Q., Wang, H., and Lin, Z. (2018).
\newblock Complete, minimal and continuous error models for the kinematic
  calibration of parallel manipulators based on poe formula.
\newblock \emph{Mechanism and Machine Theory} 121, 844--856
\bibAnnoteFile{POE}

\bibitem[{Cherubini et~al.(2010)Cherubini, Giannone, Iocchi, Nardi, and
  Palamara}]{cherubini2010policy}
Cherubini, A., Giannone, F., Iocchi, L., Nardi, D., and Palamara, P.~F. (2010).
\newblock Policy gradient learning for quadruped soccer robots.
\newblock \emph{Robotics and Autonomous Systems} 58, 872--878
\bibAnnoteFile{cherubini2010policy}

\bibitem[{Chiu et~al.(2022)Chiu, Sleiman, Mittal, Farshidian, and
  Hutter}]{chiu2022collision}
Chiu, J.-R., Sleiman, J.-P., Mittal, M., Farshidian, F., and Hutter, M. (2022).
\newblock A collision-free mpc for whole-body dynamic locomotion and
  manipulation.
\newblock In \emph{2022 International Conference on Robotics and Automation
  (ICRA)} (IEEE), 4686--4693
\bibAnnoteFile{chiu2022collision}

\bibitem[{Choromanski et~al.(2020)Choromanski, Pacchiano, Parker-Holder, Tang,
  Jain, Yang et~al.}]{choromanski2020provably}
Choromanski, K., Pacchiano, A., Parker-Holder, J., Tang, Y., Jain, D., Yang,
  Y., et~al. (2020).
\newblock Provably robust blackbox optimization for reinforcement learning.
\newblock In \emph{Conference on Robot Learning} (PMLR), 683--696
\bibAnnoteFile{choromanski2020provably}

\bibitem[{Christensen et~al.(2016)Christensen, Suresh, Hahm, and
  Cutkosky}]{christensen2016let}
Christensen, D.~L., Suresh, S.~A., Hahm, K., and Cutkosky, M.~R. (2016).
\newblock Let’s all pull together: Principles for sharing large loads in
  microrobot teams.
\newblock \emph{IEEE Robotics and Automation Letters} 1, 1089--1096
\bibAnnoteFile{christensen2016let}

\bibitem[{Cordes et~al.(2011)Cordes, Ahrns, Bartsch, Birnschein, Dettmann,
  Estable et~al.}]{cordes2011lunares}
Cordes, F., Ahrns, I., Bartsch, S., Birnschein, T., Dettmann, A., Estable, S.,
  et~al. (2011).
\newblock Lunares: Lunar crater exploration with heterogeneous multi robot
  systems.
\newblock \emph{Intelligent Service Robotics} 4, 61--89
\bibAnnoteFile{cordes2011lunares}

\bibitem[{Cordes et~al.(2012)Cordes, Roehr, Kirchner, and
  Bremen}]{cordes2012rimres}
Cordes, F., Roehr, T.~M., Kirchner, F., and Bremen, D. R. I.~C. (2012).
\newblock Rimres: A modular reconfigurable heterogeneous multi-robot
  exploration system.
\newblock In \emph{International Symposium on Artificial Intelligence, Robotics
  and Automatoin in Space}
\bibAnnoteFile{cordes2012rimres}

\bibitem[{Corke(2007)}]{DH}
Corke, P.~I. (2007).
\newblock A simple and systematic approach to assigning denavit--hartenberg
  parameters.
\newblock \emph{IEEE transactions on robotics} 23, 590--594
\bibAnnoteFile{DH}

\bibitem[{da~Silva et~al.(2021)da~Silva, Perico, Homem, and
  da~Costa~Bianchi}]{da2021deep}
da~Silva, I.~J., Perico, D.~H., Homem, T. P.~D., and da~Costa~Bianchi, R.~A.
  (2021).
\newblock Deep reinforcement learning for a humanoid robot soccer player.
\newblock \emph{Journal of Intelligent \& Robotic Systems} 102, 1--14
\bibAnnoteFile{da2021deep}

\bibitem[{Deng et~al.(2018)Deng, Xin, Zhong, and Mistry}]{deng2018object}
Deng, H., Xin, G., Zhong, G., and Mistry, M. (2018).
\newblock Object carrying of hexapod robots with integrated mechanism of leg
  and arm.
\newblock \emph{Robotics and Computer-Integrated Manufacturing} 54, 145--155
\bibAnnoteFile{deng2018object}

\bibitem[{Ding and Yang(2016)}]{ding_yang_2016}
Ding, X. and Yang, F. (2016).
\newblock Study on hexapod robot manipulation using legs.
\newblock \emph{Robotica} 34, 468–481.
\newblock \doi{10.1017/S0263574714001799}
\bibAnnoteFile{ding_yang_2016}

\bibitem[{EngineeringToolBox(2004)}]{engineeringtoolbox}
[Dataset] EngineeringToolBox (2004).
\newblock Friction - friction coefficients and calculator.
\newblock
  \url{https://www.engineeringtoolbox.com/friction-coefficients-d_778.html}
\bibAnnoteFile{engineeringtoolbox}

\bibitem[{Ewen et~al.(2021)Ewen, Sleiman, Chen, Lu, Hutter, and
  Vasudevan}]{ewen2021generating}
Ewen, P., Sleiman, J.-P., Chen, Y., Lu, W.-C., Hutter, M., and Vasudevan, R.
  (2021).
\newblock Generating continuous motion and force plans in real-time for legged
  mobile manipulation.
\newblock In \emph{2021 IEEE International Conference on Robotics and
  Automation (ICRA)} (IEEE), 4933--4939
\bibAnnoteFile{ewen2021generating}

\bibitem[{Fang and Gao(2018)}]{fang2018type}
Fang, L. and Gao, F. (2018).
\newblock Type design and behavior control for six legged robots.
\newblock \emph{Chinese Journal of Mechanical Engineering} 31, 1--12
\bibAnnoteFile{fang2018type}

\bibitem[{Ferrolho et~al.(2022)Ferrolho, Ivan, Merkt, Havoutis, and
  Vijayakumar}]{ferrolho2022roloma}
Ferrolho, H., Ivan, V., Merkt, W., Havoutis, I., and Vijayakumar, S. (2022).
\newblock Roloma: Robust loco-manipulation for quadruped robots with arms.
\newblock \emph{arXiv preprint arXiv:2203.01446}
\bibAnnoteFile{ferrolho2022roloma}

\bibitem[{Ferrolho et~al.(2020)Ferrolho, Merkt, Ivan, Wolfslag, and
  Vijayakumar}]{ferrolho2020optimizing}
Ferrolho, H., Merkt, W., Ivan, V., Wolfslag, W., and Vijayakumar, S. (2020).
\newblock Optimizing dynamic trajectories for robustness to disturbances using
  polytopic projections.
\newblock In \emph{2020 IEEE/RSJ International Conference on Intelligent Robots
  and Systems (IROS)} (IEEE), 7477--7484
\bibAnnoteFile{ferrolho2020optimizing}

\bibitem[{Flemmer and Flemmer(2014)}]{flemmer2014humanoid}
Flemmer, R. and Flemmer, C. (2014).
\newblock \emph{A humanoid robot for research into kicking rugby balls}.
\newblock Tech. rep., Research Report, 1-14, available at: http://www. massey.
  ac. nz/\~{} rcflemme/A~…
\bibAnnoteFile{flemmer2014humanoid}

\bibitem[{Friedmann et~al.(2008)Friedmann, Kiener, Petters, Thomas, Von~Stryk,
  and Sakamoto}]{friedmann2008versatile}
Friedmann, M., Kiener, J., Petters, S., Thomas, D., Von~Stryk, O., and
  Sakamoto, H. (2008).
\newblock Versatile, high-quality motions and behavior control of a humanoid
  soccer robot.
\newblock \emph{International Journal of Humanoid Robotics} 5, 417--436
\bibAnnoteFile{friedmann2008versatile}

\bibitem[{Fu et~al.(2022)Fu, Cheng, and Pathak}]{fu2022deep}
Fu, Z., Cheng, X., and Pathak, D. (2022).
\newblock Deep whole-body control: Learning a unified policy for manipulation
  and locomotion.
\newblock In \emph{Conference on Robot Learning ({CoRL})}
\bibAnnoteFile{fu2022deep}

\bibitem[{Fujita and Kimura(1998)}]{fujita1998tight}
Fujita, T. and Kimura, H. (1998).
\newblock Tight cooperative working system by multiple robots.
\newblock In \emph{IROS}. 1405--1410
\bibAnnoteFile{fujita1998tight}

\bibitem[{Gams et~al.(2022)Gams, Petri{\v{c}}, Nemec, and
  Ude}]{gams2022manipulation}
Gams, A., Petri{\v{c}}, T., Nemec, B., and Ude, A. (2022).
\newblock Manipulation learning on humanoid robots.
\newblock \emph{Current Robotics Reports} , 1--13
\bibAnnoteFile{gams2022manipulation}

\bibitem[{Gams et~al.(2015)Gams, Ude, and Morimoto}]{gams2015accelerating}
Gams, A., Ude, A., and Morimoto, J. (2015).
\newblock Accelerating synchronization of movement primitives: Dual-arm
  discrete-periodic motion of a humanoid robot.
\newblock In \emph{2015 IEEE/RSJ International Conference on Intelligent Robots
  and Systems (IROS)} (IEEE), 2754--2760
\bibAnnoteFile{gams2015accelerating}

\bibitem[{Gong et~al.(2022)Gong, Behr, Graf, Chen, Gong, and
  Daltorio}]{YifengGong}
Gong, Y., Behr, A., Graf, N., Chen, K., Gong, Z., and Daltorio, K. (2022).
\newblock {A Walking Claw for Tethered Object Retrieval}.
\newblock \emph{Journal of Mechanisms and Robotics} ,
  1--15\doi{10.1115/1.4055812}
\bibAnnoteFile{YifengGong}

\bibitem[{Gong et~al.(2023)Gong, Grezmak, Zhou, Graf, Gong, Carmichael
  et~al.}]{gong_grezmak_zhou_graf_gong_carmichael_foss_daltorio_cliffton_2023}
Gong, Y., Grezmak, J., Zhou, J., Graf, N., Gong, Z., Carmichael, N., et~al.
  (2023).
\newblock Surf zone exploration with crab-like legged robots.
\newblock \emph{IEEE International Conference on Robotics and Automation}
\bibAnnoteFile{gong_grezmak_zhou_graf_gong_carmichael_foss_daltorio_cliffton_2023}

\bibitem[{Govindaraj et~al.(2021)Govindaraj, Brinkmann, Colmenero, Nieto, But,
  de~Benedetti et~al.}]{govindaraj2021building}
Govindaraj, S., Brinkmann, W., Colmenero, F., Nieto, I.~S., But, A.,
  de~Benedetti, M., et~al. (2021).
\newblock Building a lunar infrastructure with the help of a heterogeneous
  (semi) autonomous multi-robot-team.
\newblock In \emph{proceedings of the 72nd International Astronautical Congress
  2021 (IAC-2021}
\bibAnnoteFile{govindaraj2021building}

\bibitem[{Grieco et~al.(1998)Grieco, Prieto, Armada, and Gonzalez~de
  Santos}]{728488}
Grieco, J., Prieto, M., Armada, M., and Gonzalez~de Santos, P. (1998).
\newblock A six-legged climbing robot for high payloads.
\newblock In \emph{Proceedings of the 1998 IEEE International Conference on
  Control Applications (Cat. No.98CH36104)}. vol.~1, 446--450 vol.1.
\newblock \doi{10.1109/CCA.1998.728488}
\bibAnnoteFile{728488}

\bibitem[{Hamed et~al.(2019)Hamed, Ma, Kamidi, and
  Ames}]{hamed2019hierarchical}
Hamed, K.~A., Ma, W.-L., Kamidi, V.~R., and Ames, A.~D. (2019).
\newblock Hierarchical feedback control for complex hybrid models of multiagent
  legged robotic systems.
\newblock \emph{arXiv} 2019
\bibAnnoteFile{hamed2019hierarchical}

\bibitem[{Hara et~al.(1999)Hara, Fukuda, Nishibayashi, Aiyama, Ota, and
  Arai}]{hara1999motion}
Hara, M., Fukuda, M., Nishibayashi, H., Aiyama, Y., Ota, J., and Arai, T.
  (1999).
\newblock Motion control of cooperative transportation system by quadruped
  robots based on vibration model in walking.
\newblock In \emph{Proceedings 1999 IEEE/RSJ International Conference on
  Intelligent Robots and Systems. Human and Environment Friendly Robots with
  High Intelligence and Emotional Quotients (Cat. No. 99CH36289)} (IEEE),
  vol.~3, 1651--1656
\bibAnnoteFile{hara1999motion}

\bibitem[{Hawley and Suleiman(2019{\natexlab{a}})}]{hawley2019control}
Hawley, L. and Suleiman, W. (2019{\natexlab{a}}).
\newblock Control framework for cooperative object transportation by two
  humanoid robots.
\newblock \emph{Robotics and Autonomous Systems} 115, 1--16
\bibAnnoteFile{hawley2019control}

\bibitem[{Hawley and Suleiman(2019{\natexlab{b}})}]{hawley:hal-02047495}
Hawley, L. and Suleiman, W. (2019{\natexlab{b}}).
\newblock {Control framework for cooperative object transportation by two
  humanoid robots}.
\newblock \emph{{Robotics and Autonomous Systems}} 115, 1--16
\bibAnnoteFile{hawley:hal-02047495}

\bibitem[{Heppner et~al.(2015)Heppner, Roennau, Oberl{\"a}nder, Klemm, and
  Dillmann}]{heppner2015laurope}
Heppner, G., Roennau, A., Oberl{\"a}nder, J., Klemm, S., and Dillmann, R.
  (2015).
\newblock Laurope-six legged walking robot for planetary exploration
  participating in the spacebot cup.
\newblock \emph{WS on Advanced Space Technologies for Robotics and Automation}
  2, 69--76
\bibAnnoteFile{heppner2015laurope}

\bibitem[{Hirose et~al.(2005)Hirose, Yokota, Torii, Ogata, Suganuma, Takita
  et~al.}]{hirose2005quadruped}
Hirose, S., Yokota, S., Torii, A., Ogata, M., Suganuma, S., Takita, K., et~al.
  (2005).
\newblock Quadruped walking robot centered demining system-development of
  titan-ix and its operation.
\newblock In \emph{Proceedings of the 2005 IEEE International Conference on
  Robotics and Automation} (IEEE), 1284--1290
\bibAnnoteFile{hirose2005quadruped}

\bibitem[{Hong et~al.(2022)Hong, Um, Park, and Park}]{hong2022agile}
Hong, S., Um, Y., Park, J., and Park, H.-W. (2022).
\newblock Agile and versatile climbing on ferromagnetic surfaces with a
  quadrupedal robot.
\newblock \emph{Science Robotics} 7, eadd1017
\bibAnnoteFile{hong2022agile}

\bibitem[{Hooke et~al.(2013)Hooke, Mart{\'\i}n~Duque, and
  Pedraza~Gilsanz}]{LandReview}
Hooke, R.~L., Mart{\'\i}n~Duque, J.~F., and Pedraza~Gilsanz, J.~d. (2013).
\newblock Land transformation by humans: a review.
\newblock \emph{Ene} 12, 43
\bibAnnoteFile{LandReview}

\bibitem[{Hu and Gu(2001)}]{hu2001reactive}
Hu, H. and Gu, D. (2001).
\newblock Reactive behaviours and agent architecture for sony legged robots to
  play football.
\newblock \emph{Industrial Robot: An International Journal} 28, 45--54
\bibAnnoteFile{hu2001reactive}

\bibitem[{Inoue et~al.(2002)Inoue, Nishihama, Arai, and Mae}]{1013568}
Inoue, K., Nishihama, Y., Arai, T., and Mae, Y. (2002).
\newblock Mobile manipulation of humanoid robots-body and leg control for dual
  arm manipulation.
\newblock In \emph{Proceedings 2002 IEEE International Conference on Robotics
  and Automation (Cat. No.02CH37292)}. vol.~3, 2259--2264 vol.3.
\newblock \doi{10.1109/ROBOT.2002.1013568}
\bibAnnoteFile{1013568}

\bibitem[{Inoue et~al.(2010)Inoue, Ooe, and Lee}]{5509220}
Inoue, K., Ooe, K., and Lee, S. (2010).
\newblock Pushing methods for working six-legged robots capable of locomotion
  and manipulation in three modes.
\newblock In \emph{2010 IEEE International Conference on Robotics and
  Automation}. 4742--4748.
\newblock \doi{10.1109/ROBOT.2010.5509220}
\bibAnnoteFile{5509220}

\bibitem[{Inoue et~al.(2003)Inoue, Tohge, and Iba}]{inoue2003cooperative}
Inoue, Y., Tohge, T., and Iba, H. (2003).
\newblock Cooperative transportation by humanoid robots: Learning to correct
  positioning.
\newblock In \emph{HIS} (Citeseer), 1124--1134
\bibAnnoteFile{inoue2003cooperative}

\bibitem[{Irawan et~al.(2016)Irawan, Yin, Lezaini, Razali, and
  Yusof}]{Reconfigurable}
Irawan, A., Yin, T.~Y., Lezaini, W. M. N.~W., Razali, A.~R., and Yusof, M.
  S.~M. (2016).
\newblock Reconfigurable foot-to-gripper leg for underwater bottom operator,
  hexaquad.
\newblock In \emph{2016 IEEE International Conference on Underwater System
  Technology: Theory and Applications (USYS)} (IEEE), 94--99
\bibAnnoteFile{Reconfigurable}

\bibitem[{Ji et~al.(2022)Ji, Li, Sun, Peng, Levine, Berseth
  et~al.}]{Ji2022HierarchicalRL}
Ji, Y., Li, Z., Sun, Y., Peng, X.~B., Levine, S., Berseth, G., et~al. (2022).
\newblock Hierarchical reinforcement learning for precise soccer shooting
  skills using a quadrupedal robot.
\newblock \emph{ArXiv} abs/2208.01160
\bibAnnoteFile{Ji2022HierarchicalRL}

\bibitem[{Jouandeau and Hugel(2014)}]{Jouandeau2014OptimizationOP}
Jouandeau, N. and Hugel, V. (2014).
\newblock Optimization of parametrised kicking motion for humanoid soccer
  player.
\newblock \emph{2014 IEEE International Conference on Autonomous Robot Systems
  and Competitions (ICARSC)} , 241--246
\bibAnnoteFile{Jouandeau2014OptimizationOP}

\bibitem[{Kalouche et~al.(2015)Kalouche, Rollinson, and
  Choset}]{kalouche2015modularity}
Kalouche, S., Rollinson, D., and Choset, H. (2015).
\newblock Modularity for maximum mobility and manipulation: Control of a
  reconfigurable legged robot with series-elastic actuators.
\newblock In \emph{2015 IEEE International Symposium on Safety, Security, and
  Rescue Robotics (SSRR)} (IEEE), 1--8
\bibAnnoteFile{kalouche2015modularity}

\bibitem[{Kiener and Von~Stryk(2007)}]{kiener2007cooperation}
Kiener, J. and Von~Stryk, O. (2007).
\newblock Cooperation of heterogeneous, autonomous robots: A case study of
  humanoid and wheeled robots.
\newblock In \emph{2007 IEEE/RSJ International Conference on Intelligent Robots
  and Systems} (IEEE), 959--964
\bibAnnoteFile{kiener2007cooperation}

\bibitem[{Kim and Jun(2014)}]{kim2014design}
Kim, J.-Y. and Jun, B.-H. (2014).
\newblock Design of six-legged walking robot, little crabster for underwater
  walking and operation.
\newblock \emph{Advanced Robotics} 28, 77--89
\bibAnnoteFile{kim2014design}

\bibitem[{Kim et~al.(2021)Kim, Spieler, Lupu, Ramezani, and
  Chung}]{kim2021bipedal}
Kim, K., Spieler, P., Lupu, E.-S., Ramezani, A., and Chung, S.-J. (2021).
\newblock A bipedal walking robot that can fly, slackline, and skateboard.
\newblock \emph{Science Robotics} 6, eabf8136
\bibAnnoteFile{kim2021bipedal}

\bibitem[{Kim et~al.(1999)Kim, Kang, Lee, Chung, Cho, and
  Lee}]{kim1999development}
Kim, M., Kang, S., Lee, S., Chung, W., Cho, K., and Lee, C.-W. (1999).
\newblock Development of a humanoid robot centaur-design, human interface,
  planning and control of its upper-body.
\newblock In \emph{IEEE SMC'99 Conference Proceedings. 1999 IEEE International
  Conference on Systems, Man, and Cybernetics (Cat. No. 99CH37028)} (IEEE),
  vol.~4, 948--953
\bibAnnoteFile{kim1999development}

\bibitem[{Klein et~al.(2016)Klein, Hart, and Quinn}]{klein2016autonomous}
Klein, M.~A., Hart, C., and Quinn, R.~D. (2016).
\newblock Autonomous, precision snow removal in an urban environment, featuring
  ultra-wideband beacons.
\newblock In \emph{Proceedings of the 29th International Technical Meeting of
  the Satellite Division of The Institute of Navigation (ION GNSS+ 2016)}.
  1178--1183
\bibAnnoteFile{klein2016autonomous}

\bibitem[{Koyachi et~al.(2002)Koyachi, Adachi, Izumi, and
  Hirose}]{koyachi2002control}
Koyachi, N., Adachi, H., Izumi, M., and Hirose, T. (2002).
\newblock Control of walk and manipulation by a hexapod with integrated limb
  mechanism: Melmantis-1.
\newblock In \emph{Proceedings 2002 IEEE International Conference on Robotics
  and Automation (Cat. No. 02CH37292)} (IEEE), vol.~4, 3553--3558
\bibAnnoteFile{koyachi2002control}

\bibitem[{Koyachi et~al.(1993)Koyachi, Arai, Adachi, and
  Itoh}]{koyachi1993integrated}
Koyachi, N., Arai, T., Adachi, H., and Itoh, Y. (1993).
\newblock Integrated limb mechanism of manipulation and locomotion for
  dismantling robot-basic concept for control and mechanism.
\newblock In \emph{Proceedings of 1993 IEEE/Tsukuba International Workshop on
  Advanced Robotics} (IEEE), 81--84
\bibAnnoteFile{koyachi1993integrated}

\bibitem[{Koyachi et~al.(1997)Koyachi, Arai, Adachi, Murakami, and
  Kawai}]{Melmantis}
Koyachi, N., Arai, T., Adachi, H., Murakami, A., and Kawai, K. (1997).
\newblock Mechanical design of hexapods with integrated limb mechanism:
  Melmantis-1 and melmantis-2.
\newblock In \emph{1997 8th International Conference on Advanced Robotics.
  Proceedings. ICAR'97} (IEEE), 273--278
\bibAnnoteFile{Melmantis}

\bibitem[{Kuli{\'c} et~al.(2012)Kuli{\'c}, Ott, Lee, Ishikawa, and
  Nakamura}]{kulic2012incremental}
Kuli{\'c}, D., Ott, C., Lee, D., Ishikawa, J., and Nakamura, Y. (2012).
\newblock Incremental learning of full body motion primitives and their
  sequencing through human motion observation.
\newblock \emph{The International Journal of Robotics Research} 31, 330--345
\bibAnnoteFile{kulic2012incremental}

\bibitem[{Lim et~al.(2008)Lim, Kang, Lee, Kim, and You}]{lim2008multiple}
Lim, H., Kang, Y., Lee, J., Kim, J., and You, B.-J. (2008).
\newblock Multiple humanoid cooperative control system for heterogeneous
  humanoid team.
\newblock In \emph{RO-MAN 2008-The 17th IEEE International Symposium on Robot
  and Human Interactive Communication} (IEEE), 231--236
\bibAnnoteFile{lim2008multiple}

\bibitem[{Linghan et~al.(2022)Linghan, Cong, Shengguo, Qifeng, and
  Baoqi}]{linghan2022dogget}
Linghan, M., Cong, W., Shengguo, C., Qifeng, Z., and Baoqi, Z. (2022).
\newblock Dogget: A legged manipulation system in human environments.
\newblock In \emph{2022 12th International Conference on CYBER Technology in
  Automation, Control, and Intelligent Systems (CYBER)} (IEEE), 1130--1135
\bibAnnoteFile{linghan2022dogget}

\bibitem[{Liu et~al.(2020)Liu, Iacoponi, Laschi, Wen, and
  Calisti}]{UnderwaterManipulation}
Liu, J., Iacoponi, S., Laschi, C., Wen, L., and Calisti, M. (2020).
\newblock Underwater mobile manipulation: A soft arm on a benthic legged robot.
\newblock \emph{IEEE Robotics \& Automation Magazine} 27, 12--26.
\newblock \doi{10.1109/MRA.2020.3024001}
\bibAnnoteFile{UnderwaterManipulation}

\bibitem[{Liu et~al.(2007)Liu, Tan, and Zhao}]{liu2007legged}
Liu, J., Tan, M., and Zhao, X. (2007).
\newblock Legged robots—an overview.
\newblock \emph{Transactions of the Institute of Measurement and Control} 29,
  185--202
\bibAnnoteFile{liu2007legged}

\bibitem[{Liu and Liu(2009)}]{liu2009modeling}
Liu, Y. and Liu, G. (2009).
\newblock Modeling of tracked mobile manipulators with consideration of
  track--terrain and vehicle--manipulator interactions.
\newblock \emph{Robotics and Autonomous Systems} 57, 1065--1074
\bibAnnoteFile{liu2009modeling}

\bibitem[{Lo et~al.(2017)Lo, Yang, Lin, Chu, and Lin}]{lo2017model}
Lo, A. K.-Y., Yang, Y.-H., Lin, T.-C., Chu, C.-W., and Lin, P.-C. (2017).
\newblock Model-based design and evaluation of a brachiating monkey robot with
  an active waist.
\newblock \emph{Applied Sciences} 7, 947
\bibAnnoteFile{lo2017model}

\bibitem[{Lu et~al.(2020)Lu, Tam, and Kottege}]{lu2020autonomous}
Lu, B., Tam, B., and Kottege, N. (2020).
\newblock Autonomous obstacle legipulation with a hexapod robot.
\newblock \emph{arXiv preprint arXiv:2011.06227}
\bibAnnoteFile{lu2020autonomous}

\bibitem[{Lynch and Park(2017)}]{Lynch2017ModernRM}
Lynch, K.~M. and Park, F.~C. (2017).
\newblock Modern robotics: Mechanics, planning, and control.
\newblock 196--199
\bibAnnoteFile{Lynch2017ModernRM}

\bibitem[{Lynch and Park(2018)}]{lynch_park}
[Dataset] Lynch, K.~M. and Park, F.~C. (2018).
\newblock
  \url{https://modernrobotics.northwestern.edu/nu-gm-book-resource/grasping-and-manipulation/}
\bibAnnoteFile{lynch_park}

\bibitem[{Ma et~al.(2022)Ma, Farshidian, Miki, Lee, and
  Hutter}]{ma2022combining}
Ma, Y., Farshidian, F., Miki, T., Lee, J., and Hutter, M. (2022).
\newblock Combining learning-based locomotion policy with model-based
  manipulation for legged mobile manipulators.
\newblock \emph{IEEE Robotics and Automation Letters} 7, 2377--2384
\bibAnnoteFile{ma2022combining}

\bibitem[{Maniatopoulos et~al.(2016)Maniatopoulos, Schillinger, Pong, Conner,
  and Kress-Gazit}]{maniatopoulos2016reactive}
Maniatopoulos, S., Schillinger, P., Pong, V., Conner, D.~C., and Kress-Gazit,
  H. (2016).
\newblock Reactive high-level behavior synthesis for an atlas humanoid robot.
\newblock In \emph{2016 IEEE International Conference on Robotics and
  Automation (ICRA)} (IEEE), 4192--4199
\bibAnnoteFile{maniatopoulos2016reactive}

\bibitem[{Mataric et~al.(1995)Mataric, Nilsson, and
  Simsarin}]{mataric1995cooperative}
Mataric, M.~J., Nilsson, M., and Simsarin, K.~T. (1995).
\newblock Cooperative multi-robot box-pushing.
\newblock In \emph{Proceedings 1995 IEEE/RSJ International Conference on
  Intelligent Robots and Systems. Human Robot Interaction and Cooperative
  Robots} (IEEE), vol.~3, 556--561
\bibAnnoteFile{mataric1995cooperative}

\bibitem[{Mazzolai et~al.(2012)Mazzolai, Margheri, Cianchetti, Dario, and
  Laschi}]{mazzolai2012soft}
Mazzolai, B., Margheri, L., Cianchetti, M., Dario, P., and Laschi, C. (2012).
\newblock Soft-robotic arm inspired by the octopus: Ii. from artificial
  requirements to innovative technological solutions.
\newblock \emph{Bioinspiration \& biomimetics} 7, 025005
\bibAnnoteFile{mazzolai2012soft}

\bibitem[{McGhee and Frank(1968)}]{mcghee1968stability}
McGhee, R.~B. and Frank, A.~A. (1968).
\newblock On the stability properties of quadruped creeping gaits.
\newblock \emph{Mathematical Biosciences} 3, 331--351
\bibAnnoteFile{mcghee1968stability}

\bibitem[{McGill and Lee(2011)}]{mcgill2011cooperative}
McGill, S.~G. and Lee, D.~D. (2011).
\newblock Cooperative humanoid stretcher manipulation and locomotion.
\newblock In \emph{2011 11th IEEE-RAS International Conference on Humanoid
  Robots} (IEEE), 429--433
\bibAnnoteFile{mcgill2011cooperative}

\bibitem[{Miki et~al.(2022)Miki, Lee, Hwangbo, Wellhausen, Koltun, and
  Hutter}]{miki2022learning}
Miki, T., Lee, J., Hwangbo, J., Wellhausen, L., Koltun, V., and Hutter, M.
  (2022).
\newblock Learning robust perceptive locomotion for quadrupedal robots in the
  wild.
\newblock \emph{Science Robotics} 7, eabk2822
\bibAnnoteFile{miki2022learning}

\bibitem[{Mittal et~al.(2022)Mittal, Hoeller, Farshidian, Hutter, and
  Garg}]{mittal2022articulated}
Mittal, M., Hoeller, D., Farshidian, F., Hutter, M., and Garg, A. (2022).
\newblock Articulated object interaction in unknown scenes with whole-body
  mobile manipulation.
\newblock In \emph{2022 IEEE/RSJ International Conference on Intelligent Robots
  and Systems (IROS)} (IEEE), 1647--1654
\bibAnnoteFile{mittal2022articulated}

\bibitem[{Morlando et~al.(2022)Morlando, Selvaggio, and
  Ruggiero}]{morlando2022nonprehensile}
Morlando, V., Selvaggio, M., and Ruggiero, F. (2022).
\newblock Nonprehensile object transportation with a legged manipulator.
\newblock In \emph{2022 International Conference on Robotics and Automation
  (ICRA)} (IEEE), 6628--6634
\bibAnnoteFile{morlando2022nonprehensile}

\bibitem[{Muratore et~al.(2021)Muratore, Gruner, Wiese, Belousov, Gienger, and
  Peters}]{Muratore2021NeuralPD}
Muratore, F., Gruner, T., Wiese, F., Belousov, B., Gienger, M., and Peters, J.
  (2021).
\newblock Neural posterior domain randomization.
\newblock In \emph{CoRL}
\bibAnnoteFile{Muratore2021NeuralPD}

\bibitem[{Murphy et~al.(2012)Murphy, Stephens, Abe, and Rizzi}]{murphy2012high}
Murphy, M.~P., Stephens, B., Abe, Y., and Rizzi, A.~A. (2012).
\newblock High degree-of-freedom dynamic manipulation.
\newblock In \emph{Unmanned Systems Technology XIV} (SPIE), vol. 8387, 339--348
\bibAnnoteFile{murphy2012high}

\bibitem[{Nagabandi et~al.(2018)Nagabandi, Yang, Asmar, Pandya, Kahn, Levine
  et~al.}]{nagabandi2018learning}
Nagabandi, A., Yang, G., Asmar, T., Pandya, R., Kahn, G., Levine, S., et~al.
  (2018).
\newblock Learning image-conditioned dynamics models for control of
  underactuated legged millirobots.
\newblock In \emph{2018 IEEE/RSJ International Conference on Intelligent Robots
  and Systems (IROS)} (IEEE), 4606--4613
\bibAnnoteFile{nagabandi2018learning}

\bibitem[{Nishihama et~al.(2003)Nishihama, Inoue, Arai, and Mae}]{1248924}
Nishihama, Y., Inoue, K., Arai, T., and Mae, Y. (2003).
\newblock Mobile manipulation of humanoid robots -control method for accurate
  manipulation.
\newblock In \emph{Proceedings 2003 IEEE/RSJ International Conference on
  Intelligent Robots and Systems (IROS 2003) (Cat. No.03CH37453)}. vol.~2,
  1914--1919 vol.2.
\newblock \doi{10.1109/IROS.2003.1248924}
\bibAnnoteFile{1248924}

\bibitem[{Peng et~al.(2017)Peng, Berseth, Yin, and Van
  De~Panne}]{peng2017deeploco}
Peng, X.~B., Berseth, G., Yin, K., and Van De~Panne, M. (2017).
\newblock Deeploco: Dynamic locomotion skills using hierarchical deep
  reinforcement learning.
\newblock \emph{ACM Transactions on Graphics (TOG)} 36, 1--13
\bibAnnoteFile{peng2017deeploco}

\bibitem[{Polverini et~al.(2020)Polverini, Laurenzi, Hoffman, Ruscelli, and
  Tsagarakis}]{polverini2020multi}
Polverini, M.~P., Laurenzi, A., Hoffman, E.~M., Ruscelli, F., and Tsagarakis,
  N.~G. (2020).
\newblock Multi-contact heavy object pushing with a centaur-type humanoid
  robot: Planning and control for a real demonstrator.
\newblock \emph{IEEE Robotics and Automation Letters} 5, 859--866
\bibAnnoteFile{polverini2020multi}

\bibitem[{Prattichizzo and Trinkle(2016)}]{prattichizzo2016grasping}
Prattichizzo, D. and Trinkle, J.~C. (2016).
\newblock Grasping.
\newblock In \emph{Springer handbook of robotics} (Springer). 955--988
\bibAnnoteFile{prattichizzo2016grasping}

\bibitem[{Raibert et~al.(2008)Raibert, Blankespoor, Nelson, and
  Playter}]{Bigdog}
Raibert, M., Blankespoor, K., Nelson, G., and Playter, R. (2008).
\newblock Bigdog, the rough-terrain quadruped robot.
\newblock \emph{IFAC Proceedings Volumes} 41, 10822--10825
\bibAnnoteFile{Bigdog}

\bibitem[{Rehman et~al.(2018)Rehman, Caldwell, and Semini}]{rehman2018centaur}
Rehman, B.~U., Caldwell, D.~G., and Semini, C. (2018).
\newblock Centaur robots-a survey.
\newblock In \emph{Human-Centric Robotics: Proceedings of CLAWAR 2017: 20th
  International Conference on Climbing and Walking Robots and the Support
  Technologies for Mobile Machines} (World Scientific), 247--258
\bibAnnoteFile{rehman2018centaur}

\bibitem[{Rehman et~al.(2016)Rehman, Focchi, Lee, Dallali, Caldwell, and
  Semini}]{rehman2016towards}
Rehman, B.~U., Focchi, M., Lee, J., Dallali, H., Caldwell, D.~G., and Semini,
  C. (2016).
\newblock Towards a multi-legged mobile manipulator.
\newblock In \emph{2016 IEEE International Conference on Robotics and
  Automation (ICRA)} (IEEE), 3618--3624
\bibAnnoteFile{rehman2016towards}

\bibitem[{Rigo et~al.(2022)Rigo, Chen, Gupta, and Nguyen}]{rigo2022contact}
Rigo, A., Chen, Y., Gupta, S.~K., and Nguyen, Q. (2022).
\newblock Contact optimization for non-prehensile loco-manipulation via
  hierarchical model predictive control.
\newblock \emph{arXiv preprint arXiv:2210.03442}
\bibAnnoteFile{rigo2022contact}

\bibitem[{Rober(2021)}]{rober2021worlds}
[Dataset] Rober, M. (2021).
\newblock World's longest field goal- robot vs nfl kicker.
\newblock \url{https://www.youtube.com/watch?v=P_6my53IlxY}
\bibAnnoteFile{rober2021worlds}

\bibitem[{Roennau et~al.(2014)Roennau, Heppner, Nowicki, and
  Dillmann}]{roennau2014lauron}
Roennau, A., Heppner, G., Nowicki, M., and Dillmann, R. (2014).
\newblock Lauron v: A versatile six-legged walking robot with advanced
  maneuverability.
\newblock In \emph{2014 IEEE/ASME International Conference on Advanced
  Intelligent Mechatronics} (IEEE), 82--87
\bibAnnoteFile{roennau2014lauron}

\bibitem[{Ruggiero et~al.(2018)Ruggiero, Lippiello, and
  Siciliano}]{ruggiero2018nonprehensile}
Ruggiero, F., Lippiello, V., and Siciliano, B. (2018).
\newblock Nonprehensile dynamic manipulation: A survey.
\newblock \emph{IEEE Robotics and Automation Letters} 3, 1711--1718
\bibAnnoteFile{ruggiero2018nonprehensile}

\bibitem[{Sartoretti et~al.(2018)Sartoretti, Shaw, Lam, Fan, Travers, and
  Choset}]{sartoretti2018central}
Sartoretti, G., Shaw, S., Lam, K., Fan, N., Travers, M., and Choset, H. (2018).
\newblock Central pattern generator with inertial feedback for stable
  locomotion and climbing in unstructured terrain.
\newblock In \emph{2018 IEEE International Conference on Robotics and
  Automation (ICRA)} (IEEE), 5769--5775
\bibAnnoteFile{sartoretti2018central}

\bibitem[{Sayyad et~al.(2007)Sayyad, Seth, and Seshu}]{sayyad2007single}
Sayyad, A., Seth, B., and Seshu, P. (2007).
\newblock Single-legged hopping robotics research—a review.
\newblock \emph{Robotica} 25, 587--613
\bibAnnoteFile{sayyad2007single}

\bibitem[{Schmid et~al.(2008)Schmid, Gorges, Goger, and
  Worn}]{schmid2008opening}
Schmid, A.~J., Gorges, N., Goger, D., and Worn, H. (2008).
\newblock Opening a door with a humanoid robot using multi-sensory tactile
  feedback.
\newblock In \emph{2008 IEEE International Conference on Robotics and
  Automation} (IEEE), 285--291
\bibAnnoteFile{schmid2008opening}

\bibitem[{Schroer et~al.(2004)Schroer, Boggess, Bachmann, Quinn, and
  Ritzmann}]{1308761}
Schroer, R., Boggess, M., Bachmann, R., Quinn, R., and Ritzmann, R. (2004).
\newblock Comparing cockroach and whegs robot body motions.
\newblock In \emph{IEEE International Conference on Robotics and Automation,
  2004. Proceedings. ICRA '04. 2004}. vol.~4, 3288--3293 Vol.4.
\newblock \doi{10.1109/ROBOT.2004.1308761}
\bibAnnoteFile{1308761}

\bibitem[{Schwarz et~al.(2017)Schwarz, Rodehutskors, Droeschel, Beul,
  Schreiber, Araslanov et~al.}]{schwarz2017nimbro}
Schwarz, M., Rodehutskors, T., Droeschel, D., Beul, M., Schreiber, M.,
  Araslanov, N., et~al. (2017).
\newblock Nimbro rescue: Solving disaster-response tasks with the mobile
  manipulation robot momaro.
\newblock \emph{Journal of Field Robotics} 34, 400--425
\bibAnnoteFile{schwarz2017nimbro}

\bibitem[{Sereinig et~al.(2020)Sereinig, Werth, and
  Faller}]{sereinig2020review}
Sereinig, M., Werth, W., and Faller, L.-M. (2020).
\newblock A review of the challenges in mobile manipulation: systems design and
  robocup challenges.
\newblock \emph{e \& i Elektrotechnik und Informationstechnik} 137, 297--308
\bibAnnoteFile{sereinig2020review}

\bibitem[{Shankhdhar et~al.(2022)Shankhdhar, Omer, Rawat, Thakur, Kataria,
  Shukla et~al.}]{shankhdhar2022dynamics}
Shankhdhar, P., Omer, A., Rawat, A.~S., Thakur, A.~K., Kataria, Y., Shukla, S.,
  et~al. (2022).
\newblock Dynamics and control of a mobile kicking robot.
\newblock In \emph{2022 IEEE India Council International Subsections Conference
  (INDISCON)} (IEEE), 1--10
\bibAnnoteFile{shankhdhar2022dynamics}

\bibitem[{Shaw and Sartoretti(2022)}]{shawkeyframe}
Shaw, S. and Sartoretti, G. (2022).
\newblock Keyframe-based cpg for stable gait design and online transitions in
  legged robots.
\newblock \emph{IEEE Conference on Decision and Control}
\bibAnnoteFile{shawkeyframe}

\bibitem[{Siegwart et~al.(2015)Siegwart, Hutter, Oettershagen, Burri,
  Gilitschenski, Galceran et~al.}]{siegwart2015legged}
Siegwart, R., Hutter, M., Oettershagen, P., Burri, M., Gilitschenski, I.,
  Galceran, E., et~al. (2015).
\newblock Legged and flying robots for disaster response.
\newblock In \emph{World Engineering Conference and Convention (WECC)}
  (ETH-Z{\"u}rich)
\bibAnnoteFile{siegwart2015legged}

\bibitem[{Silva and Machado(2012)}]{silva2012literature}
Silva, M.~F. and Machado, J.~T. (2012).
\newblock A literature review on the optimization of legged robots.
\newblock \emph{Journal of Vibration and Control} 18, 1753--1767
\bibAnnoteFile{silva2012literature}

\bibitem[{Sleiman et~al.(2021)Sleiman, Farshidian, Minniti, and
  Hutter}]{sleiman2021unified}
Sleiman, J.-P., Farshidian, F., Minniti, M.~V., and Hutter, M. (2021).
\newblock A unified mpc framework for whole-body dynamic locomotion and
  manipulation.
\newblock \emph{IEEE Robotics and Automation Letters} 6, 4688--4695
\bibAnnoteFile{sleiman2021unified}

\bibitem[{Sombolestan and Nguyen(2022)}]{sombolestan2022hierarchical}
Sombolestan, M. and Nguyen, Q. (2022).
\newblock Hierarchical adaptive loco-manipulation control for quadruped robots.
\newblock \emph{arXiv preprint arXiv:2209.13145}
\bibAnnoteFile{sombolestan2022hierarchical}

\bibitem[{Song et~al.(2016)Song, Kim, Rodrigue, Lee, Shim, Kim
  et~al.}]{song2016turtle}
Song, S.-H., Kim, M.-S., Rodrigue, H., Lee, J.-Y., Shim, J.-E., Kim, M.-C.,
  et~al. (2016).
\newblock Turtle mimetic soft robot with two swimming gaits.
\newblock \emph{Bioinspiration \& biomimetics} 11, 036010
\bibAnnoteFile{song2016turtle}

\bibitem[{Stone(2007)}]{stone2007intelligent}
Stone, P. (2007).
\newblock Intelligent autonomous robotics: A robot soccer case study.
\newblock \emph{Synthesis Lectures on Artificial Intelligence and Machine
  Learning} 1, 1--155
\bibAnnoteFile{stone2007intelligent}

\bibitem[{St{\"u}ber et~al.(2020)St{\"u}ber, Zito, and Stolkin}]{stuber2020let}
St{\"u}ber, J., Zito, C., and Stolkin, R. (2020).
\newblock Let's push things forward: A survey on robot pushing.
\newblock \emph{Frontiers in Robotics and AI} , 8
\bibAnnoteFile{stuber2020let}

\bibitem[{Suzumori and Faudzi(2018)}]{suzumori2018trends}
Suzumori, K. and Faudzi, A.~A. (2018).
\newblock Trends in hydraulic actuators and components in legged and tough
  robots: a review.
\newblock \emph{Advanced Robotics} 32, 458--476
\bibAnnoteFile{suzumori2018trends}

\bibitem[{Takahashi et~al.(2000)Takahashi, Arai, Mae, Inoue, and
  Koyachi}]{895266}
Takahashi, Y., Arai, T., Mae, Y., Inoue, K., and Koyachi, N. (2000).
\newblock Development of multi-limb robot with omnidirectional manipulability
  and mobility.
\newblock In \emph{Proceedings. 2000 IEEE/RSJ International Conference on
  Intelligent Robots and Systems (IROS 2000) (Cat. No.00CH37113)}. vol.~3,
  2012--2017 vol.3.
\newblock \doi{10.1109/IROS.2000.895266}
\bibAnnoteFile{895266}

\bibitem[{Teixeira et~al.(2020)Teixeira, Silva, Abreu, and
  Reis}]{Teixeira2020HumanoidRK}
Teixeira, H., Silva, T., Abreu, M., and Reis, L.~P. (2020).
\newblock Humanoid robot kick in motion ability for playing robotic soccer.
\newblock \emph{2020 IEEE International Conference on Autonomous Robot Systems
  and Competitions (ICARSC)} , 34--39
\bibAnnoteFile{Teixeira2020HumanoidRK}

\bibitem[{Thakar et~al.(2023)Thakar, Srinivasan, Al-Hussaini, Bhatt, Rajendran,
  Jung~Yoon et~al.}]{thakar2023survey}
Thakar, S., Srinivasan, S., Al-Hussaini, S., Bhatt, P.~M., Rajendran, P.,
  Jung~Yoon, Y., et~al. (2023).
\newblock A survey of wheeled mobile manipulation: A decision-making
  perspective.
\newblock \emph{Journal of Mechanisms and Robotics} 15, 020801
\bibAnnoteFile{thakar2023survey}

\bibitem[{Tsvetkov and Ramamoorthy(2022)}]{9982229}
Tsvetkov, Y. and Ramamoorthy, S. (2022).
\newblock A novel design and evaluation of a dactylus-equipped quadruped robot
  for mobile manipulation.
\newblock In \emph{2022 IEEE/RSJ International Conference on Intelligent Robots
  and Systems (IROS)}. 1633--1638.
\newblock \doi{10.1109/IROS47612.2022.9982229}
\bibAnnoteFile{9982229}

\bibitem[{Vahrenkamp et~al.(2009)Vahrenkamp, Berenson, Asfour, Kuffner, and
  Dillmann}]{vahrenkamp2009humanoid}
Vahrenkamp, N., Berenson, D., Asfour, T., Kuffner, J., and Dillmann, R. (2009).
\newblock Humanoid motion planning for dual-arm manipulation and re-grasping
  tasks.
\newblock In \emph{2009 IEEE/RSJ International Conference on Intelligent Robots
  and Systems} (IEEE), 2464--2470
\bibAnnoteFile{vahrenkamp2009humanoid}

\bibitem[{van Dam et~al.(2022)van Dam, Tulbure, Minniti, Abi-Farraj, and
  Hutter}]{van2022collision}
van Dam, J., Tulbure, A., Minniti, M.~V., Abi-Farraj, F., and Hutter, M.
  (2022).
\newblock Collision detection and identification for a legged manipulator.
\newblock \emph{arXiv preprint arXiv:2207.14745}
\bibAnnoteFile{van2022collision}

\bibitem[{Whitman et~al.(2017)Whitman, Su, Coros, Ansari, and
  Choset}]{whitman2017generating}
Whitman, J., Su, S., Coros, S., Ansari, A., and Choset, H. (2017).
\newblock Generating gaits for simultaneous locomotion and manipulation.
\newblock In \emph{2017 IEEE/RSJ International Conference on Intelligent Robots
  and Systems (IROS)} (IEEE), 2723--2729
\bibAnnoteFile{whitman2017generating}

\bibitem[{Wu et~al.(2009)Wu, Liu, Zhang, and Chen}]{wu2009survey}
Wu, Q., Liu, C., Zhang, J., and Chen, Q. (2009).
\newblock Survey of locomotion control of legged robots inspired by biological
  concept.
\newblock \emph{Science in China Series F: Information Sciences} 52, 1715--1729
\bibAnnoteFile{wu2009survey}

\bibitem[{Xin et~al.(2022)Xin, Zeng, and Qin}]{xin2022loco}
Xin, G., Zeng, F., and Qin, K. (2022).
\newblock Loco-manipulation control for arm-mounted quadruped robots: Dynamic
  and kinematic strategies.
\newblock \emph{Machines} 10, 719
\bibAnnoteFile{xin2022loco}

\bibitem[{Yang et~al.(2020{\natexlab{a}})Yang, Lancaster, Srinivasa, and
  Smith}]{yang2020benchmarking}
Yang, B., Lancaster, P.~E., Srinivasa, S.~S., and Smith, J.~R.
  (2020{\natexlab{a}}).
\newblock Benchmarking robot manipulation with the rubik's cube.
\newblock \emph{IEEE Robotics and Automation Letters} 5, 2094--2099
\bibAnnoteFile{yang2020benchmarking}

\bibitem[{Yang et~al.(2022)Yang, Sue, Li, Yang, Shen, Chi
  et~al.}]{Yang2022CollaborativeNA}
Yang, C., Sue, G.~N., Li, Z., Yang, L., Shen, H., Chi, Y., et~al. (2022).
\newblock Collaborative navigation and manipulation of a cable-towed load by
  multiple quadrupedal robots.
\newblock \emph{IEEE Robotics and Automation Letters} 7, 10041--10048
\bibAnnoteFile{Yang2022CollaborativeNA}

\bibitem[{Yang et~al.(2020{\natexlab{b}})Yang, Zhang, Zeng, Agrawal, and
  Sreenath}]{9341218}
Yang, C., Zhang, B., Zeng, J., Agrawal, A., and Sreenath, K.
  (2020{\natexlab{b}}).
\newblock Dynamic legged manipulation of a ball through multi-contact
  optimization.
\newblock In \emph{2020 IEEE/RSJ International Conference on Intelligent Robots
  and Systems (IROS)}. 7513--7520.
\newblock \doi{10.1109/IROS45743.2020.9341218}
\bibAnnoteFile{9341218}

\bibitem[{Yao et~al.(2022)Yao, Wan, Yang, Wang, Meng, Zhang
  et~al.}]{yao2022transferable}
Yao, Q., Wan, J., Yang, S., Wang, C., Meng, L., Zhang, Q., et~al. (2022).
\newblock A transferable legged mobile manipulation framework based on
  disturbance predictive control.
\newblock \emph{arXiv preprint arXiv:2203.03391}
\bibAnnoteFile{yao2022transferable}

\bibitem[{Yoshida et~al.(2001)Yoshida, Inoue, Arai, and Mae}]{936465}
Yoshida, H., Inoue, K., Arai, T., and Mae, Y. (2001).
\newblock Mobile manipulation of humanoid robots-a method of adjusting leg
  motion for improvement of arm's manipulability.
\newblock In \emph{2001 IEEE/ASME International Conference on Advanced
  Intelligent Mechatronics. Proceedings (Cat. No.01TH8556)}. vol.~1, 266--271
  vol.1.
\newblock \doi{10.1109/AIM.2001.936465}
\bibAnnoteFile{936465}

\bibitem[{Yoshida et~al.(2002)Yoshida, Inoue, Arai, and Mae}]{1013570}
Yoshida, H., Inoue, K., Arai, T., and Mae, Y. (2002).
\newblock Mobile manipulation of humanoid robots-optimal posture for generating
  large force based on statics.
\newblock In \emph{Proceedings 2002 IEEE International Conference on Robotics
  and Automation (Cat. No.02CH37292)}. vol.~3, 2271--2276 vol.3.
\newblock \doi{10.1109/ROBOT.2002.1013570}
\bibAnnoteFile{1013570}

\bibitem[{Zhang et~al.(2017)Zhang, Zhao, Chen, and Chen}]{zhang2017survey}
Zhang, Z., Zhao, J., Chen, H., and Chen, D. (2017).
\newblock A survey of bioinspired jumping robot: takeoff, air posture
  adjustment, and landing buffer.
\newblock \emph{Applied bionics and biomechanics} 2017
\bibAnnoteFile{zhang2017survey}

\bibitem[{Zhao et~al.(2019)Zhao, Gao, and Hu}]{zhao2019novel}
Zhao, Y., Gao, F., and Hu, Y. (2019).
\newblock Novel method for six-legged robots turning valves based on force
  sensing.
\newblock \emph{Mechanism and Machine Theory} 133, 64--83
\bibAnnoteFile{zhao2019novel}

\bibitem[{Zhou et~al.(2022)Zhou, Peers, Wan, Richardson, and
  Kanoulas}]{zhou2022teleman}
Zhou, C., Peers, C., Wan, Y., Richardson, R., and Kanoulas, D. (2022).
\newblock Teleman: Teleoperation for legged robot loco-manipulation using
  wearable imu-based motion capture.
\newblock \emph{arXiv preprint arXiv:2209.10314}
\bibAnnoteFile{zhou2022teleman}

\bibitem[{Zhou and Bi(2012)}]{zhou2012survey}
Zhou, X. and Bi, S. (2012).
\newblock A survey of bio-inspired compliant legged robot designs.
\newblock \emph{Bioinspiration \& biomimetics} 7, 041001
\bibAnnoteFile{zhou2012survey}

\bibitem[{Zimmermann et~al.(2021)Zimmermann, Poranne, and
  Coros}]{zimmermann2021go}
Zimmermann, S., Poranne, R., and Coros, S. (2021).
\newblock Go fetch!-dynamic grasps using boston dynamics spot with external
  robotic arm.
\newblock In \emph{2021 IEEE International Conference on Robotics and
  Automation (ICRA)} (IEEE), 4488--4494
\bibAnnoteFile{zimmermann2021go}

\bibitem[{Zuher and Romero(2012)}]{zuher2012recognition}
Zuher, F. and Romero, R. (2012).
\newblock Recognition of human motions for imitation and control of a humanoid
  robot.
\newblock In \emph{2012 Brazilian Robotics Symposium and Latin American
  Robotics Symposium} (IEEE), 190--195
\bibAnnoteFile{zuher2012recognition}

\end{thebibliography}

\clearpage
\appendix
\section{Appendix 1}\label{appendix1}

\begin{table}[h!]
\caption{Table of friction coefficients of the manipulators}
\label{Table of friction coefficients of the manipulators}
\begin{center}
\begin{tabular}{p{0.7cm}|p{6cm}|p{5cm}|p{4cm}}
\hline
\textbf{LBL}&\textbf{\qquad \qquad Robot Name}&\textbf{Object Radius/Leg Length $\frac{r}{l}$}&\textbf{Friction Coefficient $\mu$}\\
\hline
\makecell[c]{1}&Lauron~\citep{heppner2015laurope}&0.125&0.5\\
\hline
\makecell[c]{2}&Little Crabster~\citep{kim2014design}&0.036&0.35\\
\hline
\makecell[c]{3}&Hexaquad~\citep{Reconfigurable}&0.041&0.35\\
\hline
\makecell[c]{4}&TITAN-IX~\citep{hirose2005quadruped}&0.07&0.35\\
\hline
\makecell[c]{5}&Sea Hexapod~\citep{whitman2017generating}&0.1&0.5\\
\hline
\makecell[c]{6}&PH-Robot~\citep{deng2018object}&0.06&0.5\\
\hline
\makecell[c]{7}&quadruped~\citep{9982229}&0.08&0.5\\
\hline
\makecell[c]{8}&quadruped~\citep{9982229}&0.2&0.5\\
\hline
\makecell[c]{9}&MELMANTIS-1~\citep{koyachi2002control}&0.25&0.35\\
\hline
\makecell[c]{10}&Hexapod~\citep{booysen2015gait}&0.28&0.5\\
\hline
\makecell[c]{11}&PH-Robot~\citep{deng2018object}&0.3&0.5\\
\hline
\makecell[c]{12}&Snake Monster~\citep{whitman2017generating}&0.25&0.5\\
\hline
\makecell[c]{13}&NOROS-III~\citep{ding_yang_2016}&0.135&0.35\\
\hline
\makecell[c]{14}&Klann Robot~\citep{YifengGong}&0.8&0.5\\
\hline
\end{tabular}
\label{Table of legged robots using legs for manipulation}
\end{center}
\end{table}

For double-legged grasping, $sin(\theta)$ can be expressed by $\frac{r}{l}$ as Eq.~\ref{Double_Leg_eq4} based on geometry (Eq.~\ref{Double_Leg_eq3}).
\begin{equation}\label{Double_Leg_eq3}
r\cdot cos(\theta) = \frac{s}{2} + sin(\theta)\cdot l = \frac{l}{4} + sin(\theta)\cdot l
\end{equation}
\begin{equation}\label{Double_Leg_eq4}
sin(\theta) = \frac{-0.5+\sqrt{\frac{15}{4} \cdot (\frac{r}{l})^2 + 4\cdot (\frac{r}{l})^4}}{2+2\cdot (\frac{r}{l})^2}
\end{equation}

For whole-body grasping, $sin(\theta)$ can be expressed by $\frac{r}{l}$ as Eq.~\ref{Whole_Body_eq5} based on geometry (Eq.~\ref{Whole_Body_eq4}).
\begin{equation}\label{Whole_Body_eq4}
r\cdot cos(\theta) = d + sin(\theta)\cdot l = \frac{l}{4 cos(\alpha)} + sin(\theta)\cdot l
\end{equation}
\begin{equation}\label{Whole_Body_eq5}
sin(\theta) = \frac{-\frac{1}{2\cdot cos(\alpha)}+\sqrt{(4-\frac{1}{4\cdot cos(\alpha)^2})\cdot (\frac{r}{l})^2 + 4\cdot (\frac{r}{l})^4}}{2+2\cdot (\frac{r}{l})^2}
\end{equation}

\clearpage
\section{Appendix 2}

\begin{table*}[h]
\caption{Table of the maximum size of the robots and maximum carrying capacity.}
\label{Table of the maximum size of the robots and maximum carrying capacity}
\begin{center}
\begin{tabular}{p{0.6cm}|p{7cm}|p{2.75cm}|p{2.8cm}|p{2.4cm}}
\hline
\textbf{LBL}&\textbf{\qquad \qquad \qquad \quad Name}&\textbf{Robot size $cm^3$}&\textbf{Object size $cm^3$}&\textbf{Methods}\\
\hline
\makecell[c]{1}&Hexapod~\citep{5509220}&\makecell[c]{37500}&\makecell[c]{60000}&Push\\
\hline
\makecell[c]{2}&UniTree A1~\citep{rigo2022contact}&\makecell[c]{74400}&\makecell[c]{96000}&Push\\
\hline
\makecell[c]{3}&AIBO~\citep{cherubini2010policy}&\makecell[c]{16000}&\makecell[c]{500}&Kick\\
\hline
\makecell[c]{4}&Bruno~\citep{friedmann2008versatile}&\makecell[c]{34400}&\makecell[c]{500}&Kick\\
\hline
\makecell[c]{5}&Jr-AX~\citep{acosta2008modular}&\makecell[c]{19400}&\makecell[c]{500}&Kick\\
\hline
\makecell[c]{6}&DARwln-OP~\citep{da2021deep}&\makecell[c]{18200}&\makecell[c]{500}&Kick\\
\hline
\makecell[c]{7}&UniTree A1~\citep{Ji2022HierarchicalRL}&\makecell[c]{74400}&\makecell[c]{1800}&Kick\\
\hline
\makecell[c]{8}&Robotina~\citep{behnke2008hierarchical}&\makecell[c]{109800}&\makecell[c]{4200}&Kick\\
\hline
\makecell[c]{9}&Rudi~\citep{behnke2008hierarchical}&\makecell[c]{24000}&\makecell[c]{900}&Kick\\
\hline
\makecell[c]{10}&LauronV~\citep{heppner2015laurope}&\makecell[c]{440000}&\makecell[c]{2300}&Single leg\\
\hline
\makecell[c]{11}&Little Crabster~\citep{kim2014design}&\makecell[c]{2200000}&\makecell[c]{900}&Single leg\\
\hline
\makecell[c]{12}&Hexaquad~\citep{Reconfigurable}&\makecell[c]{162500}&\makecell[c]{60}&Single leg\\
\hline
\makecell[c]{13}&TITAN-IX~\citep{hirose2005quadruped}&\makecell[c]{880000}&\makecell[c]{500}&Single leg\\
\hline
\makecell[c]{14}&SEA Hexapood~\citep{whitman2017generating}&\makecell[c]{48000}&\makecell[c]{400}&Single leg\\
\hline
\makecell[c]{15}&PH-Robot~\citep{deng2018object}&\makecell[c]{125000}&\makecell[c]{300}&Single leg\\
\hline
\makecell[c]{16}&quadruped~\citep{9982229}&\makecell[c]{1670}&\makecell[c]{125}&Single leg\\
\hline
\makecell[c]{17}&MELMANTIS-1~\citep{koyachi2002control}&\makecell[c]{41000}&\makecell[c]{14800}&Double leg\\
\hline
\makecell[c]{18}&Hexapod~\citep{booysen2015gait}&\makecell[c]{255400}&\makecell[c]{164600}&Double leg\\
\hline
\makecell[c]{19}&PH-Robot~\citep{deng2018object}&\makecell[c]{125000}&\makecell[c]{16000}&Double leg\\
\hline
\makecell[c]{20}&SEA Hexapood~\citep{whitman2017generating}&\makecell[c]{48000}&\makecell[c]{18000}&Double leg\\
\hline
\makecell[c]{21}&NOROS-III~\citep{ding_yang_2016}&\makecell[c]{14100}&\makecell[c]{500}&Double leg\\
\hline
\makecell[c]{22}&Daisy~\citep{shawkeyframe}&\makecell[c]{500000}&\makecell[c]{22500}&Double leg\\
\hline
\makecell[c]{23}&quadruped~\citep{9982229}&\makecell[c]{1670}&\makecell[c]{225}&Double leg\\
\hline
\makecell[c]{24}&Klann Robot~\citep{YifengGong}&\makecell[c]{29600}&\makecell[c]{18000}&whole body\\
\hline
\makecell[c]{25}&BigDog~\citep{murphy2012high}&\makecell[c]{276600}&\makecell[c]{16500}&Dedicated arm\\
\hline
\makecell[c]{26}&Anymal~\citep{ferrolho2020optimizing}&\makecell[c]{100000}&\makecell[c]{2600}&Dedicated arm\\
\hline
\makecell[c]{27}&Spot~\citep{zimmermann2021go}&\makecell[c]{335500}&\makecell[c]{4000}&Dedicated arm\\
\hline
\makecell[c]{28}&UniTree A1~\citep{yao2022transferable}&\makecell[c]{74400}&\makecell[c]{2000}&Dedicated arm\\
\hline
\makecell[c]{29}&UniTree Go1~\citep{fu2022deep}&\makecell[c]{37500}&\makecell[c]{250}&Dedicated arm\\
\hline
\makecell[c]{30}&HyQ~\citep{rehman2016towards}&\makecell[c]{500000}&\makecell[c]{10000}&Dedicated arm\\
\hline
\makecell[c]{31}&Unitree Laikago~\citep{zhou2022teleman}&\makecell[c]{126500}&\makecell[c]{750}&Dedicated arm\\
\hline
\makecell[c]{32}&Unitree Laikago~\citep{zhou2022teleman}&\makecell[c]{126500}&\makecell[c]{750}&Dedicated arm\\
\hline
\makecell[c]{33}&SEA Hexapood~\citep{hara1999motion}&\makecell[c]{48000}&\makecell[c]{400}&Dedicated arm\\
\hline
\makecell[c]{34}&Two Bolide Y-01~\citep{chang2022design}&\makecell[c]{26400}&\makecell[c]{7700}&Team\\
\hline
\makecell[c]{35}&Four AIBO~\citep{hu2001reactive}&\makecell[c]{64300}&\makecell[c]{1400}&Team\\
\hline
\makecell[c]{36}&Two NAO~\citep{hawley:hal-02047495}&\makecell[c]{98200}&\makecell[c]{75400}&Team\\
\hline
\makecell[c]{37}&Two HOAP-1~\citep{inoue2003cooperative}&\makecell[c]{29900}&\makecell[c]{1500}&Team\\
\hline
\makecell[c]{38}&Two DARwln-OP~\citep{mcgill2011cooperative}&\makecell[c]{36400}&\makecell[c]{16000}&Team\\
\hline
\makecell[c]{39}&Two UniTree A1 and one Mini Cheetah~\citep{Yang2022CollaborativeNA}&\makecell[c]{187700}&\makecell[c]{125000}&Team\\
\hline
\makecell[c]{40}&Two Genghis-II~\citep{mataric1995cooperative}&\makecell[c]{13500}&\makecell[c]{63700}&Team\\
\hline
\end{tabular}
\label{Table of legged robot manipulation}
\end{center}
\end{table*}

\clearpage
\section{Appendix 3}

\begin{table*}[h]
\caption{Table of legged robots and control methods}
\label{Table of legged robots and control methods}
\begin{center}
\begin{tabular}{p{1cm}|p{6.5cm}|p{3.1cm}|p{5cm}}
\hline
\textbf{LBL}&\textbf{\qquad \qquad \qquad \quad Name}&\textbf{Manipulation methods}&\textbf{Control Method(s)}\\
\hline
\makecell[c]{1}&UniTree A1~\citep{rigo2022contact}&Push&Hierarchical MPC\\
\hline
\makecell[c]{2}&AIBO~\citep{cherubini2010policy}&Kick&Policy gradient reinforcement learning\\
\hline
\makecell[c]{3}&Jr-AX~\citep{acosta2008modular}&Kick&Hierarchical reactive behavior control\\
\hline
\makecell[c]{4}&DARwln-OP~\citep{da2021deep}&Kick&Deep reinforcement learning\\
\hline
\makecell[c]{5}&UniTree A1~\citep{Ji2022HierarchicalRL}&Kick&Hierarchical reinforcement learning\\
\hline
\makecell[c]{6}&Robotina~\citep{behnke2008hierarchical}&Kick&Hierarchical reactive control\\
\hline
\makecell[c]{7}&Rudi~\citep{behnke2008hierarchical}&Kick&Hierarchical reactive control\\
\hline
\makecell[c]{8}&LauronV~\citep{heppner2015laurope}&Single leg&Hierarchical reactive behavior control\\
\hline
\makecell[c]{9}&HEBI Hexapood~\citep{whitman2017generating}&Single and double leg&Trajectory optimization and CPG\\
\hline
\makecell[c]{10}&PH-Robot~\citep{deng2018object}&Single and double leg&Trajectory generation with mass estimation\\
\hline
\makecell[c]{11}&Hexapod~\citep{booysen2015gait}&Double leg&Gait generation with pitch detection\\
\hline
\makecell[c]{12}&NOROS-III~\citep{ding_yang_2016}&Double leg&Energy consumption optimization\\
\hline
\makecell[c]{13}&HEBI Daisy~\citep{shawkeyframe}&Double leg&Keyframe-based gait and CPG\\
\hline
\makecell[c]{14}&Klann Robot~\citep{YifengGong}&Whole body&Teleoperation\\
\hline
\makecell[c]{15}&BigDog~\citep{murphy2012high}&Dedicated arm&Trajectory optimization\\
\hline
\makecell[c]{16}&Anymal B~\citep{ferrolho2020optimizing}&Dedicated arm&Trajectory optimization\\
\hline
\makecell[c]{17}&Anymal C~\cite{ma2022combining}&Dedicated arm&Combine the leaning-based method and model-based method\\
\hline
\makecell[c]{18}&Spot~\citep{zimmermann2021go}&Dedicated arm&Trajectory optimization\\
\hline
\makecell[c]{19}&UniTree A1~\citep{yao2022transferable}&Dedicated arm&A hierarchical framework that merges a disturbance-based control with the reinforcement learning and the forward model\\
\hline
\makecell[c]{20}&UniTree Go1~\citep{fu2022deep}&Dedicated arm&Reinforcement learning method\\
\hline
\makecell[c]{21}&HyQ~\citep{rehman2016towards}&Dedicated arm&Ground reaction forces optimization\\
\hline
\makecell[c]{22}&Two Bolide Y-01~\citep{chang2022design}&Team&Fuzzy motion control\\
\hline
\makecell[c]{23}&Four AIBO~\citep{hu2001reactive}&Team&Hierarchical reactive behavior control\\
\hline
\makecell[c]{24}&Two DARwln-OP~\citep{mcgill2011cooperative}&Team&Gait synchronization\\
\hline
\makecell[c]{25}&Two UniTree A1 and one Mini Cheetah~\citep{Yang2022CollaborativeNA}&Team&Parallelized centralized trajectory optimization\\
\hline
\end{tabular}
\end{center}
\vspace{-5cm}
\end{table*}

\end{document}